\documentclass[twoside]{article}

\usepackage[accepted]{aistats2022}
\usepackage[utf8]{inputenc} 
\usepackage[T1]{fontenc}    
\usepackage{hyperref}       
\usepackage{url}            
\usepackage{booktabs}       
\usepackage{amsfonts}       
\usepackage{nicefrac}       
\usepackage{microtype}      
\usepackage{xcolor}         

\usepackage{caption}
\usepackage{subcaption}
\usepackage[pdftex]{graphicx}
\usepackage[title]{appendix}
\usepackage{xparse,mathtools}
\usepackage{amsmath}
\usepackage{morefloats}
\usepackage{amssymb}
\usepackage{enumitem}
\usepackage{wrapfig}

\usepackage{algorithm}
\usepackage[noend]{algpseudocode}
\makeatletter
\def\BState{\State\hskip-\ALG@thistlm}
\makeatother

\usepackage{amsmath,amsfonts,bm}

\def\eqref#1{equation~\ref{#1}}

\def\1{\bm{1}}

\DeclareMathAlphabet{\mathsfit}{\encodingdefault}{\sfdefault}{m}{sl}
\SetMathAlphabet{\mathsfit}{bold}{\encodingdefault}{\sfdefault}{bx}{n}

\DeclareMathOperator*{\argmax}{arg\,max}

\usepackage{pgfplots}
    \usepgfplotslibrary{colorbrewer}
    \pgfplotsset{
        cycle list/Dark2,
        cycle multiindex* list={
            mark list*\nextlist
            Dark2\nextlist
        },
    }
\pgfplotsset{compat=1.14}

\usepackage{xcolor, soul}
\sethlcolor{pink}
\usepackage{array}

\newcommand{\parsection}[1]{\textbf{#1 }}

\newcolumntype{C}[1]{>{\centering\let\newline\\\arraybackslash\hspace{0pt}}m{#1}}

\DeclarePairedDelimiterX{\rvect}[1]{[}{]}{\,\makervect{#1}\,}

\ExplSyntaxOn
\NewDocumentCommand{\makervect}{m}
 {
  \seq_set_split:Nnn \l_tmpa_seq { , } { #1 }
  \begin{matrix}
  \seq_use:Nn \l_tmpa_seq { & }
  \end{matrix}
 }
\ExplSyntaxOff

\newtheorem{theorem}{Result}

\definecolor{ForestGreen}{RGB}{34,139,34}
\definecolor{Cerulean}{rgb}{0.0, 0.48, 0.65}

\usepackage[frozencache=true,cachedir=.]{minted}
\usepackage{natbib}
\setcitestyle{authoryear,open={(},close={)}}

\begin{document}

%

%

\twocolumn[

\aistatstitle{Learning Proposals for Practical Energy-Based Regression}

\aistatsauthor{ Fredrik K. Gustafsson \And Martin Danelljan \And  Thomas B. Sch\"on }

\aistatsaddress{ Dept.\! of Information Technology\\Uppsala University, Sweden \And  Computer Vision Lab\\ETH Z\"urich, Switzerland \And Dept.\! of Information Technology\\Uppsala University, Sweden } ]

\begin{abstract}
    Energy-based models (EBMs) have experienced a resurgence within machine learning in recent years, including as a promising alternative for probabilistic regression. However, energy-based regression requires a proposal distribution to be manually designed for training, and an initial estimate has to be provided at test-time. We address both of these issues by introducing a conceptually simple method to automatically learn an effective proposal distribution, which is parameterized by a separate network head. To this end, we derive a surprising result, leading to a unified training objective that jointly minimizes the KL divergence from the proposal to the EBM, and the negative log-likelihood of the EBM. At test-time, we can then employ importance sampling with the trained proposal to efficiently evaluate the learned EBM and produce stand-alone predictions. Furthermore, we utilize our derived training objective to learn mixture density networks (MDNs) with a jointly trained energy-based teacher, consistently outperforming conventional MDN training on four real-world regression tasks within computer vision. Code is available at \url{https://github.com/fregu856/ebms_proposals}.
\end{abstract}

\section{INTRODUCTION}
\label{section:introduction}

Energy-based models (EBMs)~\citep{lecun2006tutorial} have been extensively studied within the field of machine learning in the past \citep{teh2003energy, bengio2003neural, mnih2005learning, hinton2006unsupervised, osadchy2005synergistic}. By using deep neural networks to parameterize the energy function \citep{xie2016theory}, EBMs have recently also experienced a significant resurgence. Most widely, EBMs are now employed for generative modelling tasks \citep{xie2017synthesizing, gao2018learning, xie2018learning, nijkamp2019learning, du2019implicit, Grathwohl2020Your, gao2020flow, pang2020learning, bao2020bi, du2021improved}. Recent work has further demonstrated the promise of EBMs for probabilistic regression, achieving impressive results for a variety of important low-dimensional regression tasks, including object detection, visual tracking, pose estimation, age estimation and robot policy learning \citep{danelljan2020probabilistic, gustafsson2019learning, gustafsson2020train, gustafsson2020accurate, hendriks2020deep, murphy2021implicit, florence2021implicit}. 

Probabilistic regression aims to estimate the predictive conditional distribution $p(y|x)$ of the target $y$ given the input $x$ \citep{kendall2017uncertainties, lakshminarayanan2017simple, chua2018deep, gast2018lightweight, ilg2018uncertainty, makansi2019overcoming, varamesh2020mixture}. As its primary advantage, the EBM directly represents this distribution by a neural network through a learnable energy function $f_\theta(x, y)$, as $p(y | x; \theta) = e^{f_\theta(x, y)}/Z(x,\theta)$. While this flexibility allows the EBM to learn highly complex and accurate distributions, it comes at a significant cost. Firstly, evaluating the resulting distribution $p(y | x; \theta)$ is generally intractable, as it requires the computation of the partition function $Z(x,\theta)$. This particularly imposes challenges for training the EBM, which often leads to application of Monte Carlo approximations with hand-tuned proposal distributions in order to pursue maximum likelihood-based learning. Secondly, EBMs are known to be difficult to sample from, which complicates their practical use at test-time. To produce predictions, prior work \citep{danelljan2020probabilistic, gustafsson2019learning, gustafsson2020train, gustafsson2020accurate} resort to gradient-based refinement of an initial estimate generated by a separately trained network.

\begin{figure*}[t]
    \centering%
    \includegraphics[width=0.925\textwidth]{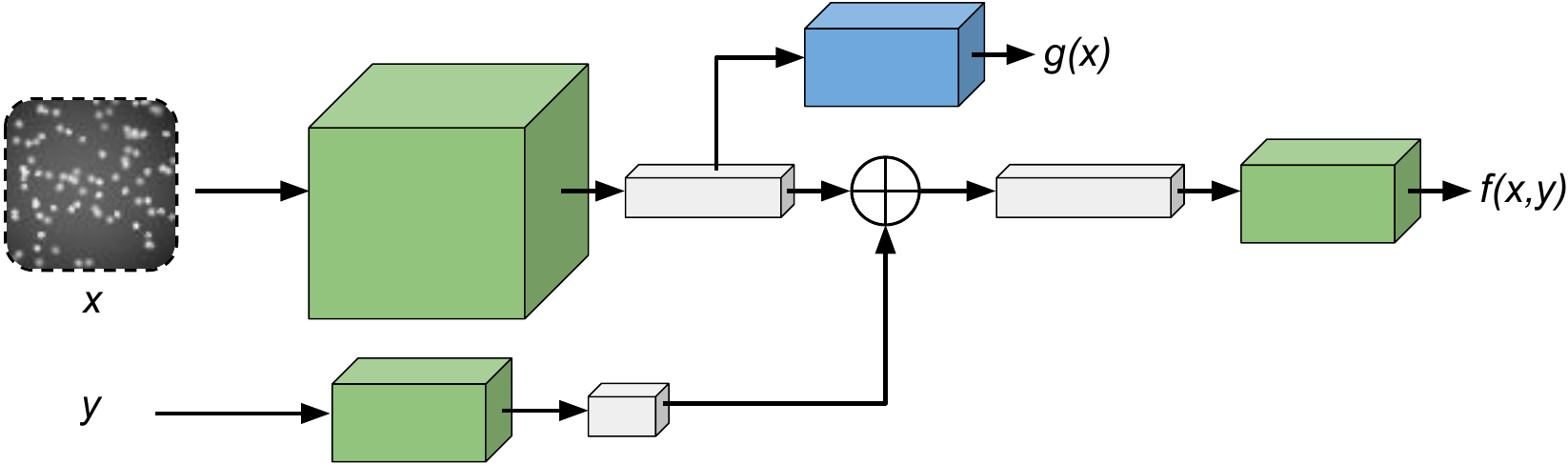}\vspace{0.5mm}
    \caption{We propose a method to automatically learn an effective MDN proposal $q(y | x; \phi)$ (\textcolor{Cerulean}{blue}) during training of the EBM $p(y | x; \theta)$ (\textcolor{ForestGreen}{green}), thus addressing the main practical limitations of energy-based regression. The MDN $q$ is trained by minimizing its KL divergence to the EBM $p$, i.e.\ by minimizing $D_\mathrm{KL}\big(p \parallel q\big)$.}\vspace{-1.0mm}
    \label{fig:overview}
\end{figure*}

In this work, we address both aforementioned drawbacks of this energy-based regression approach by jointly learning a proposal distribution $q$ during EBM training. Specifically, we parametrize the proposal using a mixture density network (MDN)~\citep{bishop1994mixture} $q(y|x;\phi)$ conditioned on the input $x$. In order to maximize its effectiveness during training, we learn $q$ by minimizing its Kullback–Leibler (KL) divergence to the EBM~$p$. To this end, we derive a surprising result, consisting of a unified objective that jointly minimizes the KL divergence from the proposal $q$ to the EBM $p$ \emph{and} the negative log-likelihood (NLL) of the latter. As our result does not rely on the reparameterization trick, it is directly applicable to a wide class of proposal distributions, including mixture models. Compared to previous approaches for training EBMs for regression, our approach does not require tedious hand-tuning of the proposal distribution, instead providing a fully learnable alternative. Moreover, rather than conditioning on the ground-truth target $y$, our proposal distribution $q$ is conditioned on the input $x$. It can therefore be employed at test-time to efficiently evaluate and sample from the EBM.

Learning the MDN $q$ according to our derived objective leads to another interesting observation: MDNs trained to mimic the EBM via this objective tend to learn more accurate predictive distributions compared to an MDN trained with the standard NLL loss. Inspired by this finding, we apply our derived result for a second purpose, namely to find a better learning formulation for MDNs. When jointly trained with an energy-based teacher network according to our objective, the resulting MDN is shown to consistently outperform the NLL baseline on challenging real-world regression tasks. In contrast to a single ground-truth sample, the EBM provides comprehensive supervision for the predictive distribution $q$, leading to a more accurate model of the underlying true distribution.

In summary, our main contributions are as follows:\vspace{-3mm}
\begin{itemize}[leftmargin=5mm]
	\setlength{\itemsep}{2pt}
	\setlength{\parskip}{-0pt}
    \item We derive an efficient and convenient objective that can be employed to train a parameterized distribution $q(y|x;\phi)$ by directly minimizing its KL divergence to a conditional EBM.
    \item We employ the proposed objective to jointly learn an effective MDN proposal distribution during EBM training, thus addressing the main practical limitations of energy-based regression.
    \item We further utilize the proposed objective to improve training of stand-alone MDNs, learning more accurate predictive distributions compared to MDNs trained by minimizing the NLL.
    \item We perform comprehensive experiments on four challenging computer vision regression tasks.
\end{itemize}
\section{BACKGROUND}
\label{section:background}

Regression entails learning to predict targets $y^\star \in \mathcal{Y}$ from inputs $x^\star \in \mathcal{X}$, given a training set of $N$ i.i.d.\ input-target pairs $\{(x_i, y_i)\}_{i=1}^{N}$, $(x_i, y_i) \sim p(x, y)$. The target space $\mathcal{Y}$ is continuous, $\mathcal{Y}=\mathbb{R}^K$ for some $K \geq 1$. We focus on probabilistic regression, which aims to not only produce a prediction $y^\star$, but also estimate the full predictive conditional distribution $p(y|x)$. This probabilistic formulation provides a more general view of the regression problem, allowing for the encapsulation of uncertainty, generation of multiple hypotheses, and handling of ill-posed settings \citep{kendall2017uncertainties, lakshminarayanan2017simple, chua2018deep, gast2018lightweight, ilg2018uncertainty, makansi2019overcoming, varamesh2020mixture}.

\subsection{Energy-Based Regression}
\label{subsection:background:eb_regression}

In energy-based regression \citep{danelljan2020probabilistic, gustafsson2019learning, gustafsson2020train}, the task is addressed by learning to model the distribution $p(y | x)$ with a conditional EBM $p(y | x; \theta)$, defined according to,
\begin{equation}
    p(y | x; \theta) = \frac{e^{f_{\theta}(x, y)}}{Z(x, \theta)}, \qquad Z(x, \theta) = \int e^{f_{\theta}(x, \tilde{y})} d\tilde{y} \,.
\label{eq:ebm_def}
\end{equation}
The EBM $p(y | x; \theta)$ is directly specified via $f_{\theta}: \mathcal{X} \times \mathcal{Y} \rightarrow \mathbb{R}$, a deep neural network (DNN) mapping any input-target pair $(x, y) \in \mathcal{X} \times \mathcal{Y}$ to a scalar $f_{\theta}(x, y) \in \mathbb{R}$. The EBM in (\ref{eq:ebm_def}) is therefore highly flexible and capable of learning complex distributions directly from data. However, the resulting distribution $p(y | x; \theta)$ is also challenging to evaluate or sample from, since its partition function $Z(x, \theta)$ generally is intractable. The EBM $p(y | x; \theta)$ is therefore quite challenging to train, and a variety of different approaches have recently been explored \citep{gustafsson2020train, song2021train}. The most straightforward approach would be to directly minimize the NLL $\mathcal{L}(\theta) =  \sum_{i=1}^{N} \log Z(x_i, \theta) - f_{\theta}(x_i, y_i)$. While exact computation of $\mathcal{L}(\theta)$ is intractable, importance sampling can be utilized to approximate the $\log Z(x_i, \theta)$ term. The DNN $f_\theta(x, y)$ can therefore be trained by minimizing the resulting loss,
\begin{equation}
    J(\theta)\!=\!\frac{1}{N}\!\sum_{i = 1}^{N} \log \bigg(\!\frac{1}{M}\!\sum_{m=1}^{M}\!\frac{e^{f_{\theta}(x_i, y_i^{(m)})}}{q(y_i^{(m)})} \bigg)\!-\!f_{\theta}(x_i, y_i)\,,
\label{eq:mle-is_loss}
\end{equation}
where $\{y_i^{(m)}\}_{m=1}^{M} \sim q(y)$ are $M$ samples drawn from a proposal distribution $q(y)$. The aforementioned approach is relatively simple, yet it has been shown effective for various regression tasks within computer vision \citep{danelljan2020probabilistic, gustafsson2019learning, gustafsson2020train}. In these works, the proposal $q(y)$ is set to a mixture of $K$ Gaussian components centered at the true target $y_i$, i.e.\ $q(y) = \frac{1}{K} \sum_{k=1}^{K} \mathcal{N}(y; y_i, \sigma_{k}^{2}I)$. Training thus requires the task-dependent hyperparameters $K$ and $\{\sigma^2_{k}\}_{k=1}^{K}$ to be carefully tuned, limiting general applicability. Moreover, this proposal $q(y)$ depends on $y_i$ and can therefore only be utilized during training. To produce a prediction $y^\star$ at test-time, previous energy-based regression methods \citep{danelljan2020probabilistic, gustafsson2019learning, gustafsson2020train, gustafsson2020accurate, hendriks2020deep, murphy2021implicit} employ gradient ascent to refine an initial estimate $\hat{y}$. This prediction strategy therefore requires access to a good initial estimate. Hence, most previous works \citep{danelljan2020probabilistic, gustafsson2019learning, gustafsson2020train, gustafsson2020accurate} even rely on a separately trained DNN to provide $\hat{y}$, further limiting general applicability.

\subsection{Mixture Density Networks}
\label{subsection:background:mdns}

Alternatively, the regression task can be addressed by learning to model the conditional distribution $p(y|x)$ with an MDN $q(y | x; \phi)$ \citep{bishop1994mixture, makansi2019overcoming, li2019generating, varamesh2020mixture}. An MDN is a mixture of $K$ components of a certain base distribution. Specifically for a Gaussian MDN, the distribution $q(y | x; \phi)$ is defined according to,
\begin{equation}
    q(y | x; \phi) = \sum_{k=1}^{K} \pi_\phi^{(k)}(x) \mathcal{N}\big(y; \mu_\phi^{(k)}(x), \sigma_\phi^{(k)}(x)I\big) \,,
\label{eq:mdn_def}
\end{equation}
where the set of Gaussian mixture parameters $\{\pi_\phi^{(k)}, \mu_\phi^{(k)}, \sigma_\phi^{(k)}\}_{k=1}^{K}$ is outputted by a DNN $g_\phi(x)$. In contrast to EBMs, the MDN distribution $q(y | x; \phi)$ is by design simple to both evaluate and sample from. The DNN $g_\phi(x)$ can thus be trained by directly minimizing the NLL $\mathcal{L}(\phi) = \sum_{i = 1}^{N} -\log q(y_i | x_i; \phi)$. While MDNs generally are less flexible models than EBMs, they are still capable of capturing multi-modality and other more complex features of the true distribution $p(y|x)$. MDNs thus offer a convenient yet quite flexible alternative to EBMs. Training an MDN $q(y | x; \phi)$ via the NLL is however known to occasionally suffer from certain inefficiencies such as mode-collapse, and various more sophisticated training methods have therefore been explored \citep{hjorth1999regularisation, rupprecht2017learning, makansi2019overcoming, cui2019multimodal, zhou2020movement}.

\section{METHOD}
\label{section:method}

We first address the main practical limitations of energy-based regression by proposing a method to automatically learn an effective proposal $q(y; \phi)$ during training of the EBM $p(y | x; \theta)$ in (\ref{eq:ebm_def}). To enable $q(y; \phi)$ to be utilized also at test-time, we condition it on the input $x$ instead of on the true target $y_i$. We further require the resulting proposal distribution $q(y | x; \phi)$ to be flexible, yet efficient and convenient to evaluate and sample from. In this work, we therefore parametrize the proposal $q(y | x; \phi)$ using an MDN. 

When training the EBM $p(y | x; \theta)$ by minimizing the approximated NLL in (\ref{eq:mle-is_loss}), we wish to use the proposal $q(y | x; \phi)$ that yields the best possible NLL approximation. In general, this is achieved when the proposal equals the EBM, i.e.\ when $q(y | x; \phi) = p(y | x; \theta)$\footnote{Details are provided in the supplementary material.}. We therefore aim to learn the proposal parameters $\phi$ by directly minimizing the KL divergence to the EBM, $D_\mathrm{KL}(p \parallel q)$. While this approach is  conceptually simple and attractive, exact computation of $D_\mathrm{KL}(p \parallel q)$ is intractable. This calls for an effective and efficient approximation, which can easily be employed during training. In Section~\ref{subsection:method:kl_approx} we show that such an approximation, interestingly enough, is achieved by simply minimizing the objective (\ref{eq:mle-is_loss}) w.r.t.\ the proposal $q(y | x; \phi)$. 

In Section~\ref{subsection:method:ebms}, we further employ this result to design a method for jointly learning the EBM $p(y | x; \theta)$ and MDN proposal $q(y | x; \phi)$. There, we also detail how $q(y | x; \phi)$ can be utilized with importance sampling to approximately evaluate and sample from the EBM at test-time. Lastly, in Section~\ref{subsection:method:mdns} we propose to utilize our derived approximation of $D_\mathrm{KL}(p \parallel q)$ as an additional loss for training MDNs. We argue that guiding an MDN towards a more flexible and accurate distribution learned by the EBM provides more extensive supervision for the MDN in a regression setting, leading to improved results.

\subsection{Learning the Proposal to Match an EBM}
\label{subsection:method:kl_approx}

We have a parameterized distribution $q(y | x; \phi)$ that we want to be a close approximation of the EBM $p(y | x; \theta)$. Specifically, we want to find the parameters $\phi$ that minimize the KL divergence between $q(y | x; \phi)$ and the EBM $p(y | x; \theta)$. Therefore, we seek to compute $\nabla_{\phi} D_\mathrm{KL}\big(p(y | x; \theta) \parallel q(y | x; \phi)\big)$, i.e.\ the gradient of the KL divergence w.r.t.\ $\phi$. The gradient $\nabla_{\phi} D_\mathrm{KL}$ is generally intractable, but can be conveniently approximated by the following result.
\begin{theorem}
    For a conditional EBM $p(y | x; \theta) = e^{f_\theta(x, y)}/\int e^{f_\theta(x, \tilde{y})} d\tilde{y}$ and distribution $q(y | x; \phi)$,
    \begin{equation}
        \nabla_{\phi} D_\mathrm{KL}\big(p \parallel q\big) \approx \nabla_{\phi} \log \bigg(\frac{1}{M} \sum_{m=1}^{M} \frac{e^{f_{\theta}(x, y^{(m)})}}{q(y^{(m)} | x; \phi)} \bigg),
    \label{eq:kl_approx}
    \end{equation}
    where $\{y^{(m)}\}_{m=1}^{M}$ are $M$ independent samples drawn from $q(y | x; \phi)$.
\label{result:kl_approx}
\end{theorem}
A complete derivation of Result~\ref{result:kl_approx} is provided in the supplementary material. 
Note that the samples $\{y^{(m)}\}_{m=1}^{M}$ in (\ref{eq:kl_approx}) are drawn from $q(y | x; \phi)$ but \emph{not} considered functions of~$\phi$, making this approximation particularly simple to compute in practice. Importantly, the approximation~(\ref{eq:kl_approx}) does \textit{not} rely on the reparameterization trick and is therefore directly applicable for a wide class of distributions $q(y | x; \phi)$, including mixture models. Given data $\{x_i\}_{i=1}^{N}$, Result~\ref{result:kl_approx} implies that $q(y | x; \phi)$ can be trained to approximate the EBM $p(y | x; \theta)$ by minimizing the loss,
\begin{equation}
    J_\mathrm{KL}(\phi) = \frac{1}{N} \sum_{i = 1}^{N} \log \bigg(\frac{1}{M} \sum_{m=1}^{M} \frac{e^{f_{\theta}(x_i, y_i^{(m)})}}{q(y_i^{(m)} | x_i; \phi)} \bigg),
\label{eq:kl_loss}
\end{equation}
where $\{y_i^{(m)}\}_{m=1}^{M} \sim q(y | x_i; \phi)$. Note that $J_\mathrm{KL}(\phi)$ is identical to the first term of the EBM loss $J(\theta)$ in (\ref{eq:mle-is_loss}). In fact, since the second term $f_{\theta}(x_i, y_i)$ in (\ref{eq:mle-is_loss}) does not depend on $\phi$, (\ref{eq:mle-is_loss}) can be used as a joint objective for training both $q(y | x; \phi)$ and the EBM $p(y | x; \theta)$.

\subsection{Practical Energy-Based Regression}
\label{subsection:method:ebms}

We first employ Result~\ref{result:kl_approx} to jointly train the EBM $p(y | x; \theta) = e^{f_\theta(x, y)}/\int e^{f_\theta(x, \tilde{y})} d\tilde{y}$ and the MDN proposal $q(y | x; \phi)$. We define the MDN $q(y | x; \phi)$ by adding a second network head $g_\phi$ onto the same backbone feature extractor shared with the EBM DNN $f_\theta$. The MDN head $g_\phi$ outputs the Gaussian mixture model parameters $\{\pi_\phi^{(k)}, \mu_\phi^{(k)}, \sigma_\phi^{(k)}\}_{k=1}^{K}$, as defined in (\ref{eq:mdn_def}). The resulting overall network architecture is illustrated in Figure~\ref{fig:overview}.

\begin{figure*}[t]
\newcommand{\wid}{3.25cm}%
\newcommand{\imwid}{0.30\textwidth}%
\centering%
			\begin{tabular}{@{\hspace{0.3cm}}c@{\hspace{3.7cm}}c@{\hspace{3.6cm}}c}
				Ground Truth &EBM &MDN Proposal
	\end{tabular}
       \includegraphics*[trim=0 0 0 36, width=\imwid]{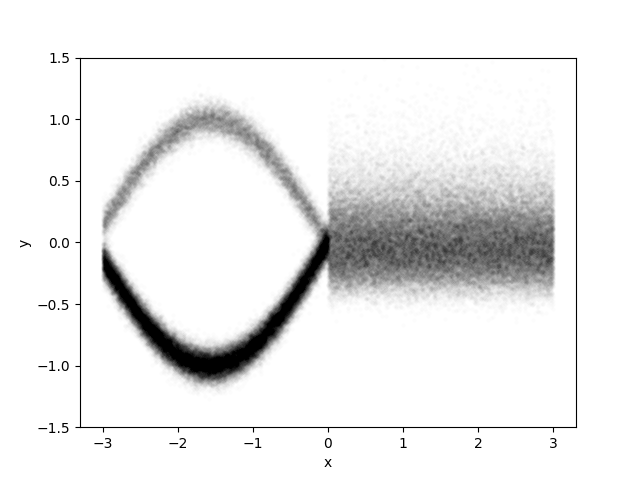}%
       \includegraphics*[trim=0 0 0 36, width=\imwid]{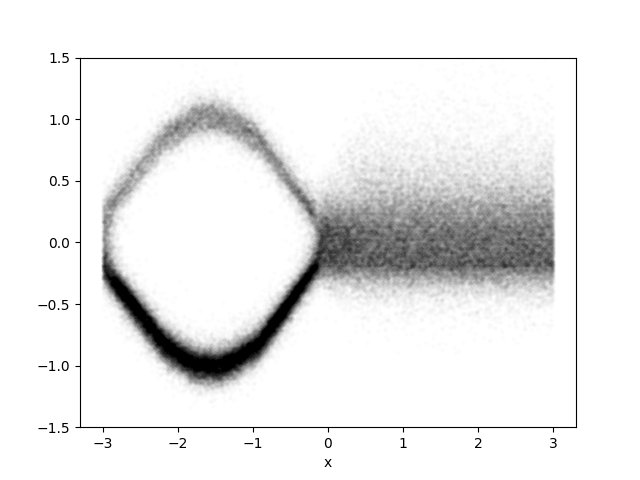}%
       \includegraphics*[trim=0 0 0 36, width=\imwid]{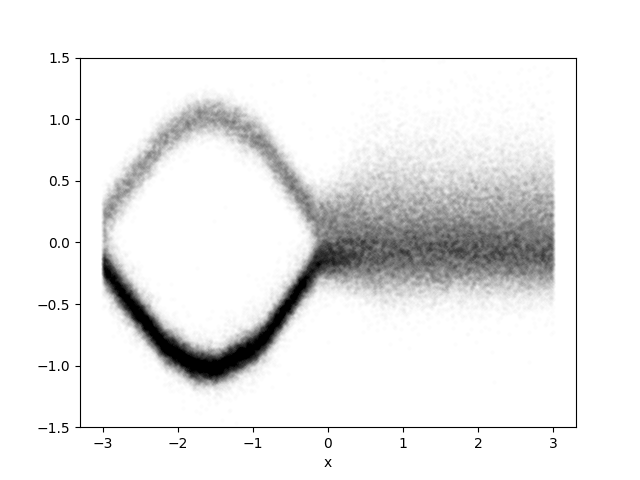}%
\caption{An illustrative 1D regression problem \citep{gustafsson2020train}, demonstrating the effectiveness of our proposed method to jointly train an EBM $p(y | x; \theta)$ and MDN proposal $q(y | x; \phi)$. In this example, the MDN has $K=4$ components. The EBM is trained using NCE with $q(y | x; \phi)$ acting as the noise distribution, whereas the MDN is trained by minimizing its KL divergence to $p(y | x; \theta)$, i.e.\ by minimizing $D_\mathrm{KL}\big(p \parallel q\big)$.}\vspace{-1.0mm}
\label{fig:illustrative}%
\end{figure*}

\subsubsection{Training}
\label{subsubsection:ebms:training}
We train the EBM $p(y | x; \theta)$ and MDN proposal $q(y | x; \phi)$ jointly using standard techniques based on stochastic gradient descent. At each iteration, we first predict the MDN mixture parameters $\{\pi_\phi^{(k)}, \mu_\phi^{(k)}, \sigma_\phi^{(k)}\}_{k=1}^{K}$ and draw $M$ samples $\{y_i^{(m)}\}_{m=1}^{M}\!\sim\!q(y | x_i; \phi)$ from the resulting distribution. The MDN parameters $\phi$ are then updated via $J_\mathrm{KL}(\phi)$ in (\ref{eq:kl_loss}), while the EBM parameters $\theta$ are updated via $J(\theta)$ in (\ref{eq:mle-is_loss}). In fact, this can be implemented by jointly minimizing (\ref{eq:mle-is_loss}) w.r.t.\ both $\theta$ and $\phi$.

EBMs can however be trained also via various alternative approaches, including noise contrastive estimation (NCE) \citep{gutmann2010noise, ma2018noise}. How EBMs should be trained specifically for regression tasks was extensively studied in \citep{gustafsson2020train}, concluding that NCE should be considered the go-to method. NCE entails training the EBM DNN $f_\theta$ by minimizing the loss,
\begin{equation}
\begin{gathered}
    J_\mathrm{NCE}(\theta) = - \frac{1}{N} \sum_{i = 1}^{N} J_\mathrm{NCE}^{(i)}(\theta),\\ 
    J_\mathrm{NCE}^{(i)}(\theta)\!=\!\log\!\frac{\exp\!\big\{\!f_{\theta}(x_i, y_i^{(0)})\!-\!\log q(y_i^{(0)})\!\big\}}{\sum\limits_{m=0}^{M}\!\exp\!\big\{\!f_{\theta}(x_i, y_i^{(m)})\!-\!\log q(y_i^{(m)})\!\big\}}, 
\label{eq:nce_loss} 
\end{gathered}
\end{equation}
where $y_i^{(0)} \triangleq y_i$, and $\{y_i^{(m)}\}_{m=1}^{M}$ are $M$ samples drawn from a noise distribution $q(y)$. The NCE loss $J_\mathrm{NCE}(\theta)$ in (\ref{eq:nce_loss}) can be interpreted as the softmax cross-entropy loss for a classification problem, distinguishing the true target $y_i$ from the $M$ noise samples $\{y_i^{(m)}\}_{m=1}^{M} \sim q(y)$. Moreover, $J_\mathrm{NCE}(\theta)$ has much similarity with the importance sampling-based loss $J(\theta)$ in (\ref{eq:mle-is_loss}) \citep{ma2018noise, jozefowicz2016exploring}. In particular, the noise distribution $q(y)$ in NCE directly corresponds to the the proposal $q$ in (\ref{eq:mle-is_loss}). In fact, all prior work \citep{gustafsson2020train, gustafsson2020accurate, hendriks2020deep} on energy-based regression using NCE has employed the same manually designed distribution $q(y) = \frac{1}{K} \sum_{k=1}^{K} \mathcal{N}(y; y_i, \sigma_{k}^{2}I)$. Due to the close relationship between NCE and importance sampling, our approach for learning the proposal distribution $q$ is also applicable for NCE-based training of the EBM. In this work, we therefore adopt the NCE loss to train the EBM $p(y | x; \theta)$, since it has been shown to achieve favorable results \citep{gustafsson2020train}.

Our approach still entails jointly training the EBM $p(y | x; \theta)$ and MDN $q(y | x; \phi)$, but employs NCE with $q(y | x; \phi)$ acting as a noise distribution for training the EBM. At each iteration we thus draw samples $\{y_i^{(m)}\}_{m=1}^{M} \sim q(y | x_i; \phi)$, update $\phi$ via the loss $J_\mathrm{KL}(\phi)$ in (\ref{eq:kl_loss}), and update $\theta$ via $J_\mathrm{NCE}(\theta)$ in (\ref{eq:nce_loss}). Note that the update of the MDN parameters $\phi$ only affects the added network head in Figure~\ref{fig:overview}, not the feature extractor. The effectiveness of this proposed joint training method is demonstrated on an illustrative 1D regression problem in Figure~\ref{fig:illustrative}. In the supplementary material (Figure~\ref{fig:illustrative_extended}), we also show an example of how both the EBM and the MDN proposal iteratively converge towards the ground truth during joint training.

Training an EBM using our joint training method is somewhat slower than using standard NCE with the manually designed $q(y) = \frac{1}{K} \sum_{k=1}^{K} \mathcal{N}(y; y_i, \sigma_{k}^{2}I)$, since we now also have to update the added network head $g_\phi$ of the MDN proposal at each iteration. For both methods, the main computational bottleneck is however the backbone feature extractor. In fact, our proposed method usually requires less total training in practice, since the task-dependent hyperparameters $K$ and $\{\sigma^2_{k}\}_{k=1}^{K}$ have to be tuned for the NCE baseline.

\subsubsection{Prediction}
To avoid evaluating the intractable $Z(x^\star, \theta)$ at test-time, previous work on energy-based regression \citep{danelljan2020probabilistic, gustafsson2019learning, gustafsson2020train, gustafsson2020accurate} approximately compute $\argmax_y p(y | x^\star; \theta) = \argmax_y f_\theta(x^\star, y)$ to produce a prediction~$y^\star$. Specifically, T steps of gradient ascent, $y \gets y + \lambda \nabla_{y} f_{\theta}(x^\star, y)$, is used to refine an initial estimate $\hat{y}$, moving it towards a local maximum of $f_\theta(x^\star, y)$. While shown to produce highly accurate predictions, this approach requires a good initial estimate $\hat{y}$ to be provided at test-time, limiting its general applicability.

\begin{figure}[t]
    \centering
    \includegraphics[width=\linewidth]{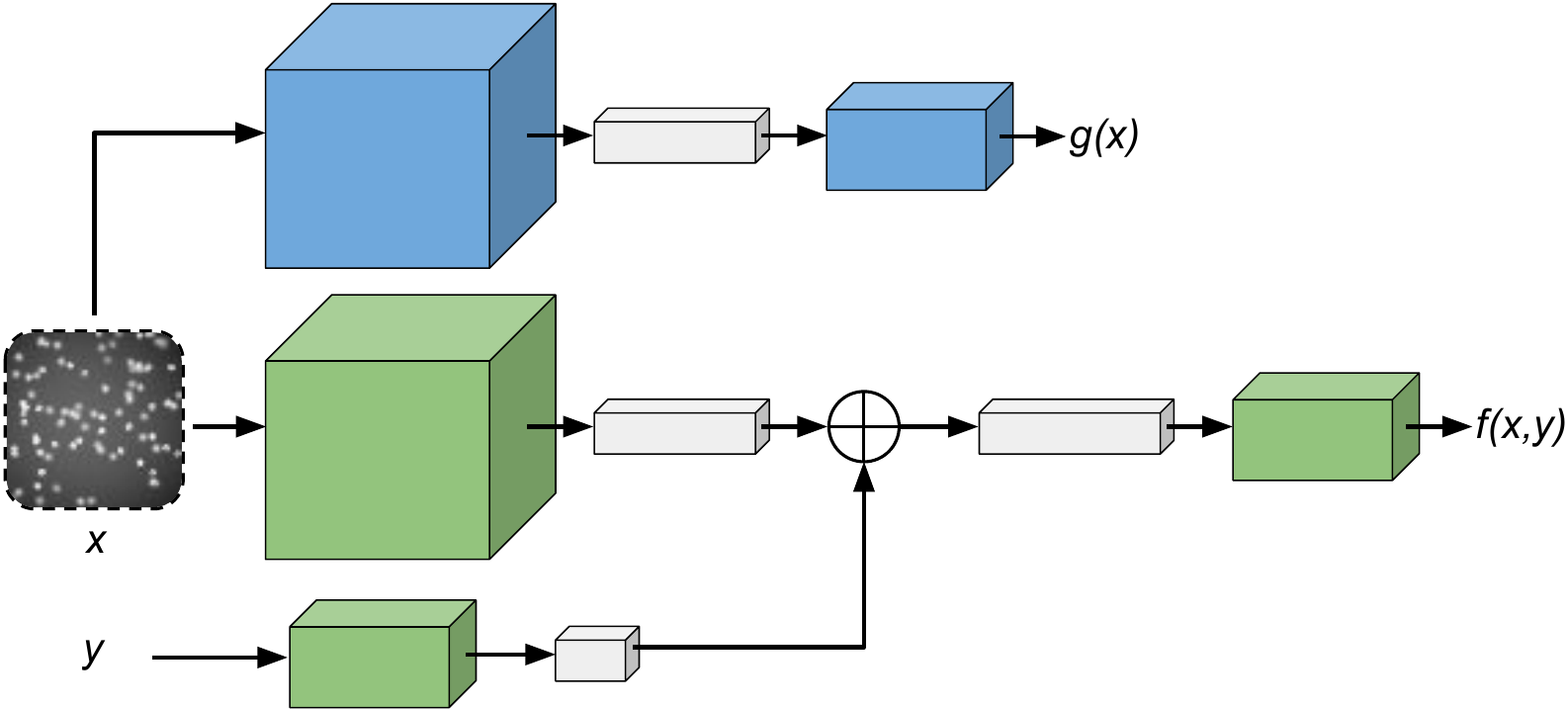}\vspace{0.5mm}
    \caption{We extend our method of jointly training an EBM $p(y | x; \theta)$ (\textcolor{ForestGreen}{green}) and MDN $q(y | x; \phi)$ (\textcolor{Cerulean}{blue}), improving MDN training. Instead of defining the MDN by adding a network head onto the EBM (Figure~\ref{fig:overview}), the MDN is now defined in terms of a full DNN $g_\phi$.}\vspace{-1.0mm}
    \label{fig:ebmdn}
\end{figure}

In contrast to previous work \citep{danelljan2020probabilistic, gustafsson2019learning, gustafsson2020train, gustafsson2020accurate}, we have access to a proposal $q(y | x; \phi)$ that is conditioned only on the input $x$ and thus can be utilized also at test-time. Since our MDN proposal $q(y | x; \phi)$ has been trained to approximate the EBM $p(y | x; \theta)$, it can be utilized with self-normalized importance sampling \citep{mcbook} to efficiently approximate expectations $\mathbb{E}_p$ w.r.t.\ the EBM $p(y | x; \theta)$,
\begin{equation}
\begin{gathered}
    \mathbb{E}_p[\xi(y)]\!=\!\int\!\xi(y) p(y | x; \theta) dy \approx \sum_{m=1}^{M} w^{(m)} \xi \big(y^{(m)}\big),\\
    w^{(m)} = \frac{e^{f_{\theta}(x, y^{(m)})}/q(y^{(m)} | x; \phi)}{\sum_{l=1}^{M} e^{f_{\theta}(x, y^{(l)})}/q(y^{(l)} | x; \phi)}. 
\label{eq:exp_importance_sampling}
\end{gathered}
\end{equation}
Here, $\{y^{(m)}\}_{m=1}^{M} \sim q(y | x; \phi)$ are samples drawn from the MDN proposal, and $\xi(y)$ is the quantity over which we are taking the expectation. For example, setting $\xi(y)\!=\!y$ in (\ref{eq:exp_importance_sampling}) enables us to approximately compute the EBM mean. In this manner, we can thus directly produce a stand-alone prediction $y^\star$ for the EBM $p(y | x; \theta)$. Using the same technique, we can also estimate the variance of the EBM as a measure of its uncertainty. 

Note that we can also draw approximate samples from the EBM $p(y | x; \theta)$ by re-sampling with replacement from the set $\{y^{(m)}\}_{m=1}^{M} \sim q(y | x; \phi)$ of proposal samples, drawing each $y^{(m)}$ with probability $w^{(m)}$ \citep{rubin1987calculation}. We demonstrate this sampling technique in Figure~\ref{fig:illustrative_sampling} in the supplementary material. There, we observe that the technique produces accurate EBM samples even when the proposal is unimodal and thus not a particularly close approximation of the EBM.

\subsection{Improved MDN Training}
\label{subsection:method:mdns}

Lastly, we employ Result~\ref{result:kl_approx} to improve the training of MDNs $q(y | x; \phi)$. We simply extend our proposed approach for jointly training an EBM $p(y | x; \theta)$ and MDN proposal $q(y | x; \phi)$ from Section~\ref{subsection:method:ebms}. Instead of defining the MDN by adding a network head onto the EBM (Figure~\ref{fig:overview}), we now define $q(y | x; \phi)$ in terms of a full DNN $g_\phi$, as illustrated in Figure~\ref{fig:ebmdn}. Now, we thus train two separate DNNs $f_\theta$ and $g_\phi$. As in Section~\ref{subsubsection:ebms:training}, we train the EBM $p(y | x; \theta)$ and MDN $q(y | x; \phi)$ jointly. At each iteration, we draw samples $\{y_i^{(m)}\}_{m=1}^{M} \sim q(y | x_i; \phi)$ and update $\theta$ via the loss $J_\mathrm{NCE}(\theta)$ in (\ref{eq:nce_loss}). The EBM is thus trained using NCE with the MDN acting as a noise distribution. At each iteration, we also update the MDN parameters $\phi$ via the loss,
\begin{equation}
\begin{gathered}
    J_\mathrm{MDN}(\phi) = \frac{1}{N} \sum_{i = 1}^{N} \frac{1}{2} \log \bigg(\frac{1}{M} \sum_{m=1}^{M} \frac{e^{f_{\theta}(x_i, y_i^{(m)})}}{q(y_i^{(m)} | x_i; \phi)} \bigg)\\
    - \frac{1}{2} \log q(y_i | x_i; \phi).
\label{eq:mdn_loss}
\end{gathered}
\end{equation}
The MDN $q(y | x; \phi)$ is thus trained by minimizing a sum of its NLL $-\log q(y_i | x_i; \phi)$ and the $J_\mathrm{KL}(\theta)$ loss in (\ref{eq:kl_loss}). Compared to conventional MDN training, we thus employ our approximation of $D_\mathrm{KL}(p \parallel q)$ as an additional loss, guiding $q(y | x; \phi)$ towards the EBM $p(y | x; \theta)$. 

In contrast to MDNs, EBMs are not restricted to distributions which are convenient to evaluate and sample. The EBM $p(y | x; \theta)$ is thus generally a more flexible model than $q(y | x; \phi)$ and therefore able to better approximate the underlying true distribution $p(y|x)$. Compared to MDNs, which define a distribution $q(y | x; \phi)$ by mapping $x$ to the set $\{\pi_\phi^{(k)}, \mu_\phi^{(k)}, \sigma_\phi^{(k)}\}_{k=1}^{K}$, the EBM $p(y | x; \theta) = e^{f_\theta(x, y)}/\int e^{f_\theta(x, \tilde{y})} d\tilde{y}$ also offers a more direct representation of the distribution via its scalar function $f_\theta(x, y)$, potentially leading to a more straightforward learning problem. Therefore, we argue that guiding the MDN $q(y | x; \phi)$ towards the EBM $p(y | x; \theta)$ during training via the loss $J_\mathrm{MDN}(\phi)$ in (\ref{eq:mdn_loss}) should help mitigate some of the known inefficiencies of MDN training. We note that our proposed joint training approach is twice as slow compared to conventional MDN training, as two separate DNNs $f_\theta$ and $g_\phi$ are updated at each iteration. After training, the EBM can however be discarded and does therefore not affect the computational cost of the MDN at test-time.
\section{RELATED WORK}
\label{section:related_work}

Our proposed approach to automatically learn a proposal during EBM training is related to the work of \citep{kim2016deep, zhai2016generative, kumar2019maximum, grathwohl2021no}, training EBMs for generative modelling tasks by jointly learning an auxiliary sampler via adversarial training. We instead train conditional EBMs for regression and are able to derive a particularly convenient KL divergence approximation (Result~\ref{result:kl_approx}). 

Our approach is also inspired by the concept of cooperative learning \citep{xie2018cooperative, xie2018cooperative_TPAMI, xie2021learning, xie2021learning_cycle}, which entails jointly training an EBM and a generator network via Markov chain Monte Carlo (MCMC) teaching. Specifically, the generator serves as a proposal and provides initial samples which are refined via MCMC to approximately sample from the EBM, training the EBM via contrastive divergence. Then, the generator network is trained to match these refined MCMC samples using a standard regression loss. Cooperative learning has recently also been extended to train EBMs for conditional generative modelling tasks \citep{xie2021cooperative, zhang2021energy}. While our proposed method also entails jointly training conditional EBMs and proposals, we specifically study the important application of low-dimensional regression. In this setting, MCMC-based training of EBMs has been shown highly inefficient \citep{gustafsson2020train}. By deriving Result~\ref{result:kl_approx}, we can instead employ the more effective training method of NCE, and train the proposal by directly minimizing its KL divergence to the EBM. Since MCMC is not employed, our proposed method is also computationally efficient, and very simple to implement, compared to cooperative learning.

\begin{table*}[t]
	\caption{Results for the EBM 1D regression experiments. Results are in terms of approximate KL divergence for the first dataset \citep{gustafsson2020train}, and in terms of approximate NLL for the second \citep{brando2019modelling}.}\vspace{-3.0mm}
	\label{tab:ebms:1dregression}
    \centering
	\resizebox{0.925\textwidth}{!}{%

\begin{tabular}{c@{\hspace{0.35cm}}|ccccc|ccc}
\toprule
&\multicolumn{5}{c|}{NCE}& \multicolumn{3}{c}{\textbf{Ours}}\\
Dataset &$\sigma_1\!=\!0.05$ &$\sigma_1\!=\!0.1$ &$\sigma_1\!=\!0.2$ &$\sigma_1\!=\!0.4$ &$\sigma_1\!=\!0.8$ &$K\!=\!1$ &$K\!=\!4$ &$K\!=\!16$\\
\midrule
\citet{gustafsson2020train}        &$0.042$ &$0.036$ &$0.040$ &$0.042$ &$0.042$ &$0.038$ &$\textbf{0.032}$ &$0.035$\\

\citet{brando2019modelling}        &$2.30$ &$1.98$ &$1.72$ &$1.67$ &$1.70$ &$1.69$ &$1.67$ &$\textbf{1.65}$\\
\bottomrule
\end{tabular}
	}\vspace{1.5mm}
\end{table*} 

\begin{table*}[t]
	\caption{Results in terms of approximate NLL for the EBM steering angle prediction experiments.}\vspace{-3.0mm}
	\label{tab:ebms:steeringangle}
    \centering
	\resizebox{0.925\textwidth}{!}{%

\begin{tabular}{ccccc|c}
\toprule
\multicolumn{5}{c|}{NCE}& \textbf{Ours}\\
$\sigma_1\!=\!0.1, \sigma_2\!=\!20$ &$\sigma_1\!=\!1, \sigma_2\!=\!20$ &$\sigma_1\!=\!2, \sigma_2\!=\!20$ &$\sigma_1\!=\!1, \sigma_2\!=\!10$ &$\sigma_1\!=\!1, \sigma_2\!=\!40$ &$K\!=\!4$\\
\midrule
$1.59\!\pm\!0.08$ &$1.51\!\pm\!0.05$ &$1.56\!\pm\!0.04$ &$2.03\!\pm\!0.14$ &$\textbf{1.39}\!\pm\!0.02$ &$1.58\!\pm\!0.13$\\
\bottomrule
\end{tabular}


	}
\end{table*} 

Our method to improve the training of an MDN by guiding it towards an EBM is related to \citep{gao2020flow}, who train a generative flow-based model jointly with an EBM through a minimax game. In contrast, our joint training method is non-adversarial and can even be implemented by directly minimizing one unified objective. On a conceptual level, our MDN training approach is also related to work on teacher-student networks and knowledge distillation \citep{hinton2015distilling, mirzadeh2020improved, xu2020knowledge, ding2021distilling}. In a knowledge distillation problem, a teacher network is utilized to improve the performance of a more lightweight student network. While knowledge distillation for regression is not a particularly well-studied topic, it has been studied for image-based regression tasks in very recent work \citep{ding2021distilling}. A student network is there enhanced by augmenting its training set with images and pseudo targets generated by a conditional GAN and a pre-trained teacher network, respectively. In contrast, our approach entails distilling the conditional EBM distribution $p(y|x; \theta)$ into a student MDN for each example in the original training set. Furthermore, our approach trains the teacher EBM and student MDN jointly, where the student MDN generates proposal samples used for training the EBM teacher.
\begin{table*}[t]
	\caption{Results in terms of approximate NLL for the EBM cell-count prediction experiments.}\vspace{-3.0mm}
	\label{tab:ebms:cells}
    \centering
	\resizebox{0.925\textwidth}{!}{%

\begin{tabular}{ccccc|c}
\toprule
\multicolumn{5}{c|}{NCE}& \textbf{Ours}\\
$\sigma_1\!=\!0.1, \sigma_2\!=\!40$ &$\sigma_1\!=\!1, \sigma_2\!=\!40$ &$\sigma_1\!=\!2, \sigma_2\!=\!40$ &$\sigma_1\!=\!1, \sigma_2\!=\!20$ &$\sigma_1\!=\!1, \sigma_2\!=\!80$ &$K\!=\!4$\\
\midrule
$2.71\!\pm\!0.07$ &$\textbf{2.64}\!\pm\!0.05$ &$2.65\!\pm\!0.05$ &$3.12\!\pm\!0.37$ &$2.70\!\pm\!0.05$ &$2.66\!\pm\!0.03$\\
\bottomrule
\end{tabular}

	}\vspace{1.5mm}
\end{table*} 

\begin{table*}[t]
	\caption{Results in terms of approximate NLL for the EBM age estimation experiments.}\vspace{-3.0mm}
	\label{tab:ebms:age}
    \centering
	\resizebox{0.925\textwidth}{!}{%

\begin{tabular}{ccccc|c}
\toprule
\multicolumn{5}{c|}{NCE}& \textbf{Ours}\\
$\sigma_1\!=\!0.01, \sigma_2\!=\!20$ &$\sigma_1\!=\!0.1, \sigma_2\!=\!20$ &$\sigma_1\!=\!1, \sigma_2\!=\!20$ &$\sigma_1\!=\!0.1, \sigma_2\!=\!10$ &$\sigma_1\!=\!0.1, \sigma_2\!=\!40$ & $K\!=\!4$\\
\midrule
$4.18\!\pm\!0.30$ &$\textbf{3.81}\!\pm\!0.18$ &$4.13\!\pm\!0.48$ &$3.97\!\pm\!0.21$ &$4.47\!\pm\!0.25$ &$4.30\!\pm\!0.30$\\
\bottomrule
\end{tabular}

	}
\end{table*}

\section{EXPERIMENTS}
\label{section:experiments}

We perform comprehensive experiments on illustrative 1D regression problems and four image-based regression tasks, which are all detailed below. We first evaluate our proposed method for automatically learning an effective proposal during EBM training in Section~\ref{subsection:experiments:ebms}. There, we compare our EBM training method with NCE, achieving highly competitive performance across all five tasks without having to tune any task-dependent hyperparameters. In Section~\ref{subsection:experiments:mdns}, we then evaluate our proposed approach for training MDNs. Compared to conventional MDN training, we consistently obtain improved test log-likelihoods. All experiments are implemented in PyTorch~\citep{paszke2019pytorch}. Example model and training code is found in the supplementary material, and our complete implementation is also made publicly available. All models were trained on individual NVIDIA TITAN Xp GPUs.

\parsection{1D Regression}
We study two illustrative 1D regression problems with $x \in \mathbb{R}$ and $y \in \mathbb{R}$. The first dataset is specified in \citep{gustafsson2020train} and contains $2\thinspace000$ training examples. It is visualized in Figure~\ref{fig:illustrative}. The second dataset is specified in \citep{brando2019modelling}, containing $1\thinspace900$ test examples and $1\thinspace700$ examples for training.

\parsection{Steering Angle Prediction}
Here, we are given an image $x$ from a forward-facing camera mounted inside of a car. The task is to predict the corresponding steering angle $y \in \mathbb{R}$ of the car at that moment. We utilize the dataset from \citep{ding2021ccgan, ding2020continuous}, containing $12\thinspace271$ examples. We randomly split the dataset into training ($80\%$) and test ($20\%$) sets. All images $x$ are of size $64 \times 64$.

\parsection{Cell-Count Prediction}
Given a synthetic fluorescence microscopy image $x$, the task is here to predict the number of cells $y \in \mathbb{R}_+$ in the image. We utilize the dataset from \citep{ding2021ccgan, ding2020continuous}, which consists of $200\thinspace000$ grayscale images of size $64 \times 64$. From this dataset, we randomly draw $10\thinspace000$ images each to construct training and test sets. An example image $x$ is visualized in Figure~\ref{fig:overview}.

\parsection{Age Estimation}
In age estimation, we are given an image $x$ of a person's face and are tasked with predicting the age $y \in \mathbb{R}_+$ of this person. We utilize the UTKFace~\citep{zhang2017age} dataset, specifically the processed version provided by \citet{ding2021ccgan, ding2020continuous}. This dataset contains $14\thinspace760$ examples, which we randomly split into training ($80\%$) and test ($20\%$) sets. All images $x$ are of size $64 \times 64$.

\parsection{Head-Pose Estimation}
In this case, we are given an image $x$ of a person, and the task is to predict the orientation $y \in \mathbb{R}^3$ of this person's head. Here, $y$ is the yaw, pitch and roll angles of the head. We utilize the BIWI~\citep{fanelli2013random} dataset, specifically the processed version provided by \citet{yang2019fsa}. We employ protocol 2 as defined in \citep{yang2019fsa}, giving $5\thinspace065$ test images and $10\thinspace613$ images for training. All images $x$ are of size $64 \times 64$.

 \begin{table*}[t]
	\caption{Results in terms of approximate NLL for the EBM head-pose estimation experiments.}\vspace{-3.0mm}
	\label{tab:ebms:headpose}
    \centering
	\resizebox{0.925\textwidth}{!}{%

\begin{tabular}{ccccc|c}
\toprule
\multicolumn{5}{c|}{NCE}& \textbf{Ours}\\
$\sigma_1\!=\!0.1, \sigma_2\!=\!20$ &$\sigma_1\!=\!1, \sigma_2\!=\!20$ &$\sigma_1\!=\!2, \sigma_2\!=\!20$ &$\sigma_1\!=\!1, \sigma_2\!=\!10$ &$\sigma_1\!=\!1, \sigma_2\!=\!40$ &$K\!=\!4$\\
\midrule
$13.68\!\pm\!0.10$ &$10.99\!\pm\!0.29$ &$10.85\!\pm\!0.11$ &$10.73\!\pm\!0.19$ &$11.20\!\pm\!0.15$ &$\textbf{9.51}\!\pm\!0.07$\\
\bottomrule
\end{tabular}

	}\vspace{1.5mm}
\end{table*} 

\begin{table*}[t]
	\caption{Results in terms of NLL for the MDN experiments on four image-based regression tasks.}\vspace{-3.0mm}
	\label{tab:mdn}
    \centering
	\resizebox{0.925\textwidth}{!}{%

\begin{tabular}{l@{\hspace{0.35cm}}|ccc|ccc}
\toprule
&\multicolumn{3}{c|}{NLL}& \multicolumn{3}{c}{\textbf{Ours}}\\
Task &$K\!=\!4$ &$K\!=\!8$ &$K\!=\!16$ &$K\!=\!4$ &$K\!=\!8$ &$K\!=\!16$\\
\midrule
Steering angle        &$1.45\!\pm\!0.13$ &$1.25\!\pm\!0.05$ &- &$\textbf{1.00}\!\pm\!0.03$ &$1.01\!\pm\!0.04$ &-\\

Cell-count            &$2.80\!\pm\!0.09$ &$2.90\!\pm\!0.06$ &- &$2.80\!\pm\!0.06$ &$\textbf{2.75}\!\pm\!0.06$ &-\\

Age                   &$4.88\!\pm\!0.21$ &$4.71\!\pm\!0.35$ &- &$\textbf{3.57}\!\pm\!0.28$ &$3.65\!\pm\!0.18$ &-\\

Head-pose             &$11.02\!\pm\!0.16$ &$10.68\!\pm\!0.39$ &$10.71\!\pm\!0.17$ &$\textbf{8.69}\!\pm\!0.10$ &$8.79\!\pm\!0.06$ &$8.77\!\pm\!0.09$\\
\bottomrule
\end{tabular}




	}
\end{table*} 

\subsection{EBM Experiments}
\label{subsection:experiments:ebms}

We first evaluate our proposed method for automatically learning an effective proposal during EBM training, by performing extensive experiments on all five regression tasks.

\parsection{1D Regression}
The EBM DNN $f_\theta(x, y)$ is here a simple feed-forward network, taking $x \in \mathbb{R}$ and $y \in \mathbb{R}$ as inputs. Separate sets of fully-connected layers extract features $h_x \in \mathbb{R}^{10}$ from $x$ and $h_y \in \mathbb{R}^{10}$ from $y$. The two feature vectors are then concatenated and processed to output $f_\theta(x, y) \in \mathbb{R}$. The MDN network head $g_\phi(x)$ takes the feature $h_x \in \mathbb{R}^{10}$ as input and outputs $\{\pi_\phi^{(k)}, \mu_\phi^{(k)}, \sigma_\phi^{(k)}\}_{k=1}^{K}$. We use the ADAM~\citep{kingma2014adam} optimizer to jointly train $f_\theta$ and $g_\phi$. For the first dataset, we follow \citep{gustafsson2020train} and evaluate the training methods in terms of how close the EBM $p(y | x; \theta)$ is to the known ground truth $p(y|x)$, as measured by the (approximately computed) KL divergence. For the second dataset~\citep{brando2019modelling} we approximately compute the test set NLL of the EBM $p(y | x; \theta)$, by evaluating $f_\theta(x, y)$ at densely sampled $y$ values in an interval $[y_{\text{min}}, y_{\text{max}}]$. We compare our proposed approach with training the EBM using NCE, employing the noise distribution $q(y) = \frac{1}{2} \sum_{k=1}^{2} \mathcal{N}(y; y_i, \sigma_{k}^{2}I)$. Following \citep{gustafsson2020train}, we set $\sigma_1 = 0.1$, $\sigma_2 = 8\sigma_1$. We also report results for the values $\sigma_1 \in \{0.05, 0.2, 0.4, 0.8\}$. For our proposed approach, we report results for using $K \in \{1, 4, 16\}$ components in the MDN proposal. We train $20$ networks for each setting and dataset, and report the mean of the $5$ best runs. The results are found in Table~\ref{tab:ebms:1dregression}. We observe that our proposed training method achieves highly competitive performance for all values of $K$. For NCE, the performance varies quite significantly with $\sigma_1$, which would have to be tuned for each dataset.

\parsection{Image-Based Regression}
We employ a virtually identical network architecture for all four image-based regression tasks, only making minor modifications for the head-pose estimation task to accommodate the higher target dimension $y \in \mathbb{R}^3$. The EBM DNN $f_\theta(x, y)$ is composed of a ResNet18~\citep{he2016deep} that extracts features $h_x \in \mathbb{R}^{512}$ from the input image $x$. From the target $y$, fully-connected layers extract features $h_y \in \mathbb{R}^{128}$. After concatenation of $h_x$ and  $h_y$, fully-connected layers then output $f_\theta(x, y) \in \mathbb{R}$. The MDN network head $g_\phi(x)$ takes the image features $h_x \in \mathbb{R}^{512}$ as input and outputs $\{\pi_\phi^{(k)}, \mu_\phi^{(k)}, \sigma_\phi^{(k)}\}_{k=1}^{K}$. Again, we use ADAM to jointly train $f_\theta$ and $g_\phi$. We evaluate the training methods by approximately computing the test set NLL of the EBM $p(y | x; \theta)$. We compare our proposed approach with training the EBM using NCE, employing the noise distribution $q(y) = \frac{1}{2} \sum_{k=1}^{2} \mathcal{N}(y; y_i, \sigma_{k}^{2}I)$. For each of the four tasks, we initially set $\{\sigma_1, \sigma_2\}$ to what was used for age estimation and head-pose estimation in \citep{gustafsson2019learning} and then carefully tune them further. Based on the 1D regression results in Table~\ref{tab:ebms:1dregression}, we use $K=4$ components in the MDN proposal for our proposed approach. We train $20$ networks for each setting and dataset, and report the mean of the $5$ best runs. The results are found in Table~\ref{tab:ebms:steeringangle} to Table~\ref{tab:ebms:headpose}. We observe that our proposed training method achieves highly competitive performance. In particular, our method significantly outperforms the NCE baseline on the more challenging head-pose estimation task (Table~\ref{tab:ebms:headpose}), which has a multi-dimensional target space. Note that we use an identical architecture for the MDN proposal in our training method across all four tasks, while the task-dependent NCE hyperparameters $\{\sigma_1, \sigma_2\}$ are tuned directly on each of the corresponding \emph{test} sets. Thus, NCE is here a very strong baseline.

\subsection{MDN Experiments}
\label{subsection:experiments:mdns}

Lastly, we perform experiments on the four image-based regression tasks to evaluate our proposed approach for training MDNs $q(y | x; \phi)$. For the EBM DNN $f_\theta(x, y)$, an identical network architecture is used as in the EBM experiments (Section~\ref{subsection:experiments:ebms}). The MDN network $g_\phi(x)$ is now a full DNN. It consists of a ResNet18 that extracts image features $h_x \in \mathbb{R}^{512}$, and a head of fully-connected layers that takes $h_x \in \mathbb{R}^{512}$ as input and outputs $\{\pi_\phi^{(k)}, \mu_\phi^{(k)}, \sigma_\phi^{(k)}\}_{k=1}^{K}$. As described in Section~\ref{subsection:method:mdns}, the MDN DNN $g_\phi$ is trained by minimizing the loss $J_{\mathrm{MDN}}(\phi)$ in (\ref{eq:mdn_loss}), whereas $f_\theta$ is trained via $J_\mathrm{NCE}(\theta)$ in (\ref{eq:nce_loss}). As in the previous experiments, ADAM is used to jointly train $f_\theta$ and $g_\phi$. We compare our proposed approach with the conventional MDN training method, i.e.\ minimizing the NLL $\sum_{i = 1}^{N} -\log q(y_i | x_i; \phi)$. We evaluate the training methods in terms of test set NLL, for MDNs with $K \in \{4, 8, 16\}$ components. We train $20$ networks for each setting and dataset, and report the mean of the $5$ best runs. The results are found in Table~\ref{tab:mdn}. We observe that our proposed training method consistently outperforms the baseline of pure NLL training. For the steering angle prediction and age estimation tasks, our approach achieves substantial improvements. Moreover, in the particularly challenging head-pose estimation task, our approach outperforms the standard MDN by a significant margin.

\section{CONCLUSION}
\label{section:conclusion}

We derived an efficient and convenient objective that can be employed to train a parameterized distribution $q(y|x;\phi)$ by minimizing its KL divergence to a conditional EBM $p(y|x; \theta)$. We then applied the derived objective to jointly learn an effective MDN proposal distribution during EBM training, thus addressing the main practical limitations of energy-based regression. We evaluated our proposed EBM training method on illustrative 1D regression problems and real-world regression tasks within computer vision, achieving highly competitive performance without having to tune any task-dependent hyperparameters. Lastly, we employed the derived objective to improve training of stand-alone MDNs, consistently obtaining more accurate predictive distributions compared to conventional MDN training. Future directions include estimating the EBM uncertainty via test-time use of the trained MDN proposal, and applying our MDN training approach to additional tasks.

\subsubsection*{Acknowledgements}
This research was supported by the Swedish Foundation for Strategic Research via the project \emph{ASSEMBLE} (contract number: RIT15-0012), by the Swedish Research Council via the project \emph{NewLEADS - New Directions in Learning Dynamical Systems} (contract number: 621-2016-06079), and by the \emph{Kjell \& M\"arta Beijer Foundation}.

\bibliography{references}

\begin{thebibliography}{66}
\providecommand{\natexlab}[1]{#1}
\providecommand{\url}[1]{\texttt{#1}}
\expandafter\ifx\csname urlstyle\endcsname\relax
  \providecommand{\doi}[1]{doi: #1}\else
  \providecommand{\doi}{doi: \begingroup \urlstyle{rm}\Url}\fi

\bibitem[Bao et~al.(2020)Bao, Li, Xu, Su, Zhu, and Zhang]{bao2020bi}
F.~Bao, C.~Li, K.~Xu, H.~Su, J.~Zhu, and B.~Zhang.
\newblock Bi-level score matching for learning energy-based latent variable
  models.
\newblock In \emph{Advances in Neural Information Processing Systems
  (NeurIPS)}, 2020.

\bibitem[Bengio et~al.(2003)Bengio, Ducharme, Vincent, and
  Jauvin]{bengio2003neural}
Y.~Bengio, R.~Ducharme, P.~Vincent, and C.~Jauvin.
\newblock A neural probabilistic language model.
\newblock \emph{Journal of machine learning research}, 3\penalty0
  (Feb):\penalty0 1137--1155, 2003.

\bibitem[Bishop(1994)]{bishop1994mixture}
C.~M. Bishop.
\newblock Mixture density networks, 1994.

\bibitem[Brando et~al.(2019)Brando, Rodr{\'\i}guez-Serrano, Vitria, and
  Rubio]{brando2019modelling}
A.~Brando, J.~A. Rodr{\'\i}guez-Serrano, J.~Vitria, and A.~Rubio.
\newblock Modelling heterogeneous distributions with an uncountable mixture of
  asymmetric laplacians.
\newblock In \emph{Advances in Neural Information Processing Systems
  (NeurIPS)}, 2019.

\bibitem[Chua et~al.(2018)Chua, Calandra, McAllister, and Levine]{chua2018deep}
K.~Chua, R.~Calandra, R.~McAllister, and S.~Levine.
\newblock Deep reinforcement learning in a handful of trials using
  probabilistic dynamics models.
\newblock In \emph{Advances in Neural Information Processing Systems
  (NeurIPS)}, pages 4759--4770, 2018.

\bibitem[Cui et~al.(2019)Cui, Radosavljevic, Chou, Lin, Nguyen, Huang,
  Schneider, and Djuric]{cui2019multimodal}
H.~Cui, V.~Radosavljevic, F.-C. Chou, T.-H. Lin, T.~Nguyen, T.-K. Huang,
  J.~Schneider, and N.~Djuric.
\newblock Multimodal trajectory predictions for autonomous driving using deep
  convolutional networks.
\newblock In \emph{International Conference on Robotics and Automation (ICRA)},
  pages 2090--2096. IEEE, 2019.

\bibitem[Danelljan et~al.(2020)Danelljan, Gool, and
  Timofte]{danelljan2020probabilistic}
M.~Danelljan, L.~V. Gool, and R.~Timofte.
\newblock Probabilistic regression for visual tracking.
\newblock In \emph{Proceedings of the IEEE/CVF Conference on Computer Vision
  and Pattern Recognition (CVPR)}, pages 7183--7192, 2020.

\bibitem[Ding et~al.(2020)Ding, Wang, Xu, Welch, and Wang]{ding2020continuous}
X.~Ding, Y.~Wang, Z.~Xu, W.~J. Welch, and Z.~J. Wang.
\newblock Continuous conditional generative adversarial networks for image
  generation: Novel losses and label input mechanisms.
\newblock \emph{arXiv preprint arXiv:2011.07466}, 2020.

\bibitem[Ding et~al.(2021{\natexlab{a}})Ding, Wang, Xu, Wang, and
  Welch]{ding2021distilling}
X.~Ding, Y.~Wang, Z.~Xu, Z.~J. Wang, and W.~J. Welch.
\newblock Distilling and transferring knowledge via c{GAN}-generated samples
  for image classification and regression.
\newblock \emph{arXiv preprint arXiv:2104.03164}, 2021{\natexlab{a}}.

\bibitem[Ding et~al.(2021{\natexlab{b}})Ding, Wang, Xu, Welch, and
  Wang]{ding2021ccgan}
X.~Ding, Y.~Wang, Z.~Xu, W.~J. Welch, and Z.~J. Wang.
\newblock Cc{GAN}: Continuous conditional generative adversarial networks for
  image generation.
\newblock In \emph{International Conference on Learning Representations
  (ICLR)}, 2021{\natexlab{b}}.
\newblock URL \url{https://openreview.net/forum?id=PrzjugOsDeE}.

\bibitem[Du and Mordatch(2019)]{du2019implicit}
Y.~Du and I.~Mordatch.
\newblock Implicit generation and modeling with energy based models.
\newblock In \emph{Advances in Neural Information Processing Systems
  (NeurIPS)}, 2019.

\bibitem[Du et~al.(2021)Du, Li, Tenenbaum, and Mordatch]{du2021improved}
Y.~Du, S.~Li, J.~Tenenbaum, and I.~Mordatch.
\newblock Improved contrastive divergence training of energy based models.
\newblock In \emph{International Conference on Machine Learning (ICML)}, 2021.

\bibitem[Fanelli et~al.(2013)Fanelli, Dantone, Gall, Fossati, and
  Van~Gool]{fanelli2013random}
G.~Fanelli, M.~Dantone, J.~Gall, A.~Fossati, and L.~Van~Gool.
\newblock Random forests for real time 3d face analysis.
\newblock \emph{International Journal of Computer Vision (IJCV)}, 101\penalty0
  (3):\penalty0 437--458, 2013.

\bibitem[Florence et~al.(2021)Florence, Lynch, Zeng, Ramirez, Wahid, Downs,
  Wong, Lee, Mordatch, and Tompson]{florence2021implicit}
P.~Florence, C.~Lynch, A.~Zeng, O.~A. Ramirez, A.~Wahid, L.~Downs, A.~Wong,
  J.~Lee, I.~Mordatch, and J.~Tompson.
\newblock Implicit behavioral cloning.
\newblock In \emph{The 5th Annual Conference on Robot Learning (CoRL)}, 2021.

\bibitem[Gao et~al.(2018)Gao, Lu, Zhou, Zhu, and Nian~Wu]{gao2018learning}
R.~Gao, Y.~Lu, J.~Zhou, S.-C. Zhu, and Y.~Nian~Wu.
\newblock Learning generative convnets via multi-grid modeling and sampling.
\newblock In \emph{Proceedings of the IEEE Conference on Computer Vision and
  Pattern Recognition (CVPR)}, pages 9155--9164, 2018.

\bibitem[Gao et~al.(2020)Gao, Nijkamp, Kingma, Xu, Dai, and Wu]{gao2020flow}
R.~Gao, E.~Nijkamp, D.~P. Kingma, Z.~Xu, A.~M. Dai, and Y.~N. Wu.
\newblock Flow contrastive estimation of energy-based models.
\newblock In \emph{Proceedings of the IEEE/CVF Conference on Computer Vision
  and Pattern Recognition (CVPR)}, pages 7518--7528, 2020.

\bibitem[Gast and Roth(2018)]{gast2018lightweight}
J.~Gast and S.~Roth.
\newblock Lightweight probabilistic deep networks.
\newblock In \emph{Proceedings of the IEEE Conference on Computer Vision and
  Pattern Recognition (CVPR)}, pages 3369--3378, 2018.

\bibitem[Grathwohl et~al.(2020)Grathwohl, Wang, Jacobsen, Duvenaud, Norouzi,
  and Swersky]{Grathwohl2020Your}
W.~Grathwohl, K.-C. Wang, J.-H. Jacobsen, D.~Duvenaud, M.~Norouzi, and
  K.~Swersky.
\newblock Your classifier is secretly an energy based model and you should
  treat it like one.
\newblock In \emph{International Conference on Learning Representations
  (ICLR)}, 2020.

\bibitem[Grathwohl et~al.(2021)Grathwohl, Kelly, Hashemi, Norouzi, Swersky, and
  Duvenaud]{grathwohl2021no}
W.~S. Grathwohl, J.~J. Kelly, M.~Hashemi, M.~Norouzi, K.~Swersky, and
  D.~Duvenaud.
\newblock No {MCMC} for me: Amortized sampling for fast and stable training of
  energy-based models.
\newblock In \emph{International Conference on Learning Representations
  (ICLR)}, 2021.

\bibitem[Gustafsson et~al.(2020{\natexlab{a}})Gustafsson, Danelljan, Bhat, and
  Sch{\"o}n]{gustafsson2019learning}
F.~K. Gustafsson, M.~Danelljan, G.~Bhat, and T.~B. Sch{\"o}n.
\newblock Energy-based models for deep probabilistic regression.
\newblock In \emph{Proceedings of the European Conference on Computer Vision
  (ECCV)}, August 2020{\natexlab{a}}.

\bibitem[Gustafsson et~al.(2020{\natexlab{b}})Gustafsson, Danelljan, Timofte,
  and Sch{\"o}n]{gustafsson2020train}
F.~K. Gustafsson, M.~Danelljan, R.~Timofte, and T.~B. Sch{\"o}n.
\newblock How to train your energy-based model for regression.
\newblock In \emph{Proceedings of the British Machine Vision Conference
  (BMVC)}, September 2020{\natexlab{b}}.

\bibitem[Gustafsson et~al.(2021)Gustafsson, Danelljan, and
  Sch{\"o}n]{gustafsson2020accurate}
F.~K. Gustafsson, M.~Danelljan, and T.~B. Sch{\"o}n.
\newblock Accurate 3{D} object detection using energy-based models.
\newblock In \emph{Proceedings of the IEEE/CVF Conference on Computer Vision
  and Pattern Recognition (CVPR) Workshops}, 2021.

\bibitem[Gutmann and Hyv{\"a}rinen(2010)]{gutmann2010noise}
M.~Gutmann and A.~Hyv{\"a}rinen.
\newblock Noise-contrastive estimation: {A} new estimation principle for
  unnormalized statistical models.
\newblock In \emph{Proceedings of the International Conference on Artificial
  Intelligence and Statistics (AISTATS)}, pages 297--304, 2010.

\bibitem[He et~al.(2016)He, Zhang, Ren, and Sun]{he2016deep}
K.~He, X.~Zhang, S.~Ren, and J.~Sun.
\newblock Deep residual learning for image recognition.
\newblock In \emph{Proceedings of the IEEE Conference on Computer Vision and
  Pattern Recognition (CVPR)}, pages 770--778, 2016.

\bibitem[Hendriks et~al.(2021)Hendriks, Gustafsson, Ribeiro, Wills, and
  Sch{\"o}n]{hendriks2020deep}
J.~N. Hendriks, F.~K. Gustafsson, A.~H. Ribeiro, A.~G. Wills, and T.~B.
  Sch{\"o}n.
\newblock Deep energy-based {NARX} models.
\newblock In \emph{Proceedings of the 19th IFAC Symposium on System
  Identification (SYSID)}, July 2021.

\bibitem[Hinton et~al.(2006)Hinton, Osindero, Welling, and
  Teh]{hinton2006unsupervised}
G.~Hinton, S.~Osindero, M.~Welling, and Y.-W. Teh.
\newblock Unsupervised discovery of nonlinear structure using contrastive
  backpropagation.
\newblock \emph{Cognitive science}, 30\penalty0 (4):\penalty0 725--731, 2006.

\bibitem[Hinton et~al.(2015)Hinton, Vinyals, and Dean]{hinton2015distilling}
G.~Hinton, O.~Vinyals, and J.~Dean.
\newblock Distilling the knowledge in a neural network.
\newblock \emph{arXiv preprint arXiv:1503.02531}, 2015.

\bibitem[Hjorth and Nabney(1999)]{hjorth1999regularisation}
L.~U. Hjorth and I.~T. Nabney.
\newblock Regularisation of mixture density networks.
\newblock In \emph{1999 Ninth International Conference on Artificial Neural
  Networks ICANN 99.(Conf. Publ. No. 470)}, volume~2, pages 521--526. IET,
  1999.

\bibitem[Ilg et~al.(2018)Ilg, Cicek, Galesso, Klein, Makansi, Hutter, and
  Bro]{ilg2018uncertainty}
E.~Ilg, O.~Cicek, S.~Galesso, A.~Klein, O.~Makansi, F.~Hutter, and T.~Bro.
\newblock Uncertainty estimates and multi-hypotheses networks for optical flow.
\newblock In \emph{Proceedings of the European Conference on Computer Vision
  (ECCV)}, pages 652--667, 2018.

\bibitem[Jozefowicz et~al.(2016)Jozefowicz, Vinyals, Schuster, Shazeer, and
  Wu]{jozefowicz2016exploring}
R.~Jozefowicz, O.~Vinyals, M.~Schuster, N.~Shazeer, and Y.~Wu.
\newblock Exploring the limits of language modeling.
\newblock \emph{arXiv preprint arXiv:1602.02410}, 2016.

\bibitem[Kendall and Gal(2017)]{kendall2017uncertainties}
A.~Kendall and Y.~Gal.
\newblock What uncertainties do we need in {B}ayesian deep learning for
  computer vision?
\newblock In \emph{Advances in Neural Information Processing Systems
  (NeurIPS)}, pages 5574--5584, 2017.

\bibitem[Kim and Bengio(2016)]{kim2016deep}
T.~Kim and Y.~Bengio.
\newblock Deep directed generative models with energy-based probability
  estimation.
\newblock \emph{arXiv preprint arXiv:1606.03439}, 2016.

\bibitem[Kingma and Ba(2014)]{kingma2014adam}
D.~P. Kingma and J.~Ba.
\newblock Adam: A method for stochastic optimization.
\newblock \emph{arXiv preprint arXiv:1412.6980}, 2014.

\bibitem[Kumar et~al.(2019)Kumar, Ozair, Goyal, Courville, and
  Bengio]{kumar2019maximum}
R.~Kumar, S.~Ozair, A.~Goyal, A.~Courville, and Y.~Bengio.
\newblock Maximum entropy generators for energy-based models.
\newblock \emph{arXiv preprint arXiv:1901.08508}, 2019.

\bibitem[Lakshminarayanan et~al.(2017)Lakshminarayanan, Pritzel, and
  Blundell]{lakshminarayanan2017simple}
B.~Lakshminarayanan, A.~Pritzel, and C.~Blundell.
\newblock Simple and scalable predictive uncertainty estimation using deep
  ensembles.
\newblock In \emph{Advances in Neural Information Processing Systems
  (NeurIPS)}, pages 6402--6413, 2017.

\bibitem[LeCun et~al.(2006)LeCun, Chopra, Hadsell, Ranzato, and
  Huang]{lecun2006tutorial}
Y.~LeCun, S.~Chopra, R.~Hadsell, M.~Ranzato, and F.~Huang.
\newblock A tutorial on energy-based learning.
\newblock \emph{Predicting structured data}, 1\penalty0 (0), 2006.

\bibitem[Li and Lee(2019)]{li2019generating}
C.~Li and G.~H. Lee.
\newblock Generating multiple hypotheses for 3d human pose estimation with
  mixture density network.
\newblock In \emph{Proceedings of the IEEE/CVF Conference on Computer Vision
  and Pattern Recognition (CVPR)}, pages 9887--9895, 2019.

\bibitem[Ma and Collins(2018)]{ma2018noise}
Z.~Ma and M.~Collins.
\newblock Noise contrastive estimation and negative sampling for conditional
  models: {C}onsistency and statistical efficiency.
\newblock In \emph{Proceedings of the Conference on Empirical Methods in
  Natural Language Processing (EMNLP)}, pages 3698--3707, 2018.

\bibitem[Makansi et~al.(2019)Makansi, Ilg, Cicek, and
  Brox]{makansi2019overcoming}
O.~Makansi, E.~Ilg, O.~Cicek, and T.~Brox.
\newblock Overcoming limitations of mixture density networks: A sampling and
  fitting framework for multimodal future prediction.
\newblock In \emph{Proceedings of the IEEE Conference on Computer Vision and
  Pattern Recognition (CVPR)}, pages 7144--7153, 2019.

\bibitem[Mirzadeh et~al.(2020)Mirzadeh, Farajtabar, Li, Levine, Matsukawa, and
  Ghasemzadeh]{mirzadeh2020improved}
S.~I. Mirzadeh, M.~Farajtabar, A.~Li, N.~Levine, A.~Matsukawa, and
  H.~Ghasemzadeh.
\newblock Improved knowledge distillation via teacher assistant.
\newblock In \emph{Proceedings of the AAAI Conference on Artificial
  Intelligence}, pages 5191--5198, 2020.

\bibitem[Mnih and Hinton(2005)]{mnih2005learning}
A.~Mnih and G.~Hinton.
\newblock Learning nonlinear constraints with contrastive backpropagation.
\newblock In \emph{Proceedings of the IEEE International Joint Conference on
  Neural Networks}, volume~2, pages 1302--1307. IEEE, 2005.

\bibitem[Murphy et~al.(2021)Murphy, Esteves, Jampani, Ramalingam, and
  Makadia]{murphy2021implicit}
K.~Murphy, C.~Esteves, V.~Jampani, S.~Ramalingam, and A.~Makadia.
\newblock Implicit-pdf: Non-parametric representation of probability
  distributions on the rotation manifold.
\newblock In \emph{International Conference on Machine Learning (ICML)}, 2021.

\bibitem[Nijkamp et~al.(2019)Nijkamp, Hill, Zhu, and Wu]{nijkamp2019learning}
E.~Nijkamp, M.~Hill, S.-C. Zhu, and Y.~N. Wu.
\newblock Learning non-convergent non-persistent short-run mcmc toward
  energy-based model.
\newblock In \emph{Advances in Neural Information Processing Systems
  (NeurIPS)}, pages 5233--5243, 2019.

\bibitem[Osadchy et~al.(2005)Osadchy, Miller, and Cun]{osadchy2005synergistic}
M.~Osadchy, M.~L. Miller, and Y.~L. Cun.
\newblock Synergistic face detection and pose estimation with energy-based
  models.
\newblock In \emph{Advances in Neural Information Processing Systems
  (NeurIPS)}, pages 1017--1024, 2005.

\bibitem[Owen(2013)]{mcbook}
A.~B. Owen.
\newblock \emph{Monte Carlo theory, methods and examples}.
\newblock 2013.

\bibitem[Pang et~al.(2020)Pang, Han, Nijkamp, Zhu, and Wu]{pang2020learning}
B.~Pang, T.~Han, E.~Nijkamp, S.-C. Zhu, and Y.~N. Wu.
\newblock Learning latent space energy-based prior model.
\newblock In \emph{Advances in Neural Information Processing Systems
  (NeurIPS)}, 2020.

\bibitem[Paszke et~al.(2019)Paszke, Gross, Massa, Lerer, Bradbury, Chanan,
  Killeen, Lin, Gimelshein, Antiga, et~al.]{paszke2019pytorch}
A.~Paszke, S.~Gross, F.~Massa, A.~Lerer, J.~Bradbury, G.~Chanan, T.~Killeen,
  Z.~Lin, N.~Gimelshein, L.~Antiga, et~al.
\newblock Py{T}orch: An imperative style, high-performance deep learning
  library.
\newblock In \emph{Advances in Neural Information Processing Systems
  (NeurIPS)}, pages 8024--8035, 2019.

\bibitem[Rubin(1987)]{rubin1987calculation}
D.~B. Rubin.
\newblock The calculation of posterior distributions by data augmentation:
  Comment: A noniterative sampling/importance resampling alternative to the
  data augmentation algorithm for creating a few imputations when fractions of
  missing information are modest: The sir algorithm.
\newblock \emph{Journal of the American Statistical Association}, 82\penalty0
  (398):\penalty0 543--546, 1987.

\bibitem[Rupprecht et~al.(2017)Rupprecht, Laina, DiPietro, Baust, Tombari,
  Navab, and Hager]{rupprecht2017learning}
C.~Rupprecht, I.~Laina, R.~DiPietro, M.~Baust, F.~Tombari, N.~Navab, and G.~D.
  Hager.
\newblock Learning in an uncertain world: Representing ambiguity through
  multiple hypotheses.
\newblock In \emph{Proceedings of the IEEE International Conference on Computer
  Vision (ICCV)}, pages 3591--3600, 2017.

\bibitem[Song and Kingma(2021)]{song2021train}
Y.~Song and D.~P. Kingma.
\newblock How to train your energy-based models.
\newblock \emph{arXiv preprint arXiv:2101.03288}, 2021.

\bibitem[Teh et~al.(2003)Teh, Welling, Osindero, and Hinton]{teh2003energy}
Y.~W. Teh, M.~Welling, S.~Osindero, and G.~E. Hinton.
\newblock Energy-based models for sparse overcomplete representations.
\newblock \emph{Journal of Machine Learning Research}, 4\penalty0
  (Dec):\penalty0 1235--1260, 2003.

\bibitem[Varamesh and Tuytelaars(2020)]{varamesh2020mixture}
A.~Varamesh and T.~Tuytelaars.
\newblock Mixture dense regression for object detection and human pose
  estimation.
\newblock In \emph{Proceedings of the IEEE/CVF Conference on Computer Vision
  and Pattern Recognition (CVPR)}, pages 13086--13095, 2020.

\bibitem[Xie et~al.(2016)Xie, Lu, Zhu, and Wu]{xie2016theory}
J.~Xie, Y.~Lu, S.-C. Zhu, and Y.~Wu.
\newblock A theory of generative convnet.
\newblock In \emph{International Conference on Machine Learning (ICML)}, pages
  2635--2644, 2016.

\bibitem[Xie et~al.(2017)Xie, Zhu, and Nian~Wu]{xie2017synthesizing}
J.~Xie, S.-C. Zhu, and Y.~Nian~Wu.
\newblock Synthesizing dynamic patterns by spatial-temporal generative convnet.
\newblock In \emph{Proceedings of the IEEE Conference on Computer Vision and
  Pattern Recognition (CVPR)}, pages 7093--7101, 2017.

\bibitem[Xie et~al.(2018{\natexlab{a}})Xie, Lu, Gao, and
  Wu]{xie2018cooperative}
J.~Xie, Y.~Lu, R.~Gao, and Y.~N. Wu.
\newblock Cooperative learning of energy-based model and latent variable model
  via {MCMC} teaching.
\newblock In \emph{Proceedings of the AAAI Conference on Artificial
  Intelligence}, volume~32, 2018{\natexlab{a}}.

\bibitem[Xie et~al.(2018{\natexlab{b}})Xie, Lu, Gao, Zhu, and
  Wu]{xie2018cooperative_TPAMI}
J.~Xie, Y.~Lu, R.~Gao, S.-C. Zhu, and Y.~N. Wu.
\newblock Cooperative training of descriptor and generator networks.
\newblock \emph{IEEE Transactions on Pattern Analysis and Machine Intelligence
  (TPAMI)}, 42\penalty0 (1):\penalty0 27--45, 2018{\natexlab{b}}.

\bibitem[Xie et~al.(2018{\natexlab{c}})Xie, Zheng, Gao, Wang, Zhu, and
  Wu]{xie2018learning}
J.~Xie, Z.~Zheng, R.~Gao, W.~Wang, S.-C. Zhu, and Y.~N. Wu.
\newblock Learning descriptor networks for 3d shape synthesis and analysis.
\newblock In \emph{Proceedings of the IEEE Conference on Computer Vision and
  Pattern Recognition (CVPR)}, pages 8629--8638, 2018{\natexlab{c}}.

\bibitem[Xie et~al.(2021{\natexlab{a}})Xie, Zheng, Fang, Zhu, and
  Wu]{xie2021cooperative}
J.~Xie, Z.~Zheng, X.~Fang, S.-C. Zhu, and Y.~N. Wu.
\newblock Cooperative training of fast thinking initializer and slow thinking
  solver for conditional learning.
\newblock \emph{IEEE Transactions on Pattern Analysis and Machine Intelligence
  (TPAMI)}, 2021{\natexlab{a}}.

\bibitem[Xie et~al.(2021{\natexlab{b}})Xie, Zheng, Fang, Zhu, and
  Wu]{xie2021learning_cycle}
J.~Xie, Z.~Zheng, X.~Fang, S.-C. Zhu, and Y.~N. Wu.
\newblock Learning cycle-consistent cooperative networks via alternating mcmc
  teaching for unsupervised cross-domain translation.
\newblock In \emph{The Thirty-Fifth AAAI Conference on Artificial Intelligence
  (AAAI)}, 2021{\natexlab{b}}.

\bibitem[Xie et~al.(2021{\natexlab{c}})Xie, Zheng, and Li]{xie2021learning}
J.~Xie, Z.~Zheng, and P.~Li.
\newblock Learning energy-based model with variational auto-encoder as
  amortized sampler.
\newblock In \emph{The Thirty-Fifth AAAI Conference on Artificial Intelligence
  (AAAI)}, volume~2, 2021{\natexlab{c}}.

\bibitem[Xu et~al.(2020)Xu, Liu, Li, and Loy]{xu2020knowledge}
G.~Xu, Z.~Liu, X.~Li, and C.~C. Loy.
\newblock Knowledge distillation meets self-supervision.
\newblock In \emph{Proceedings of the European Conference on Computer Vision
  (ECCV)}, pages 588--604, 2020.

\bibitem[Yang et~al.(2019)Yang, Chen, Lin, and Chuang]{yang2019fsa}
T.-Y. Yang, Y.-T. Chen, Y.-Y. Lin, and Y.-Y. Chuang.
\newblock {FSA-Net}: Learning fine-grained structure aggregation for head pose
  estimation from a single image.
\newblock In \emph{Proceedings of the IEEE Conference on Computer Vision and
  Pattern Recognition (CVPR)}, pages 1087--1096, 2019.

\bibitem[Zhai et~al.(2016)Zhai, Cheng, Feris, and Zhang]{zhai2016generative}
S.~Zhai, Y.~Cheng, R.~Feris, and Z.~Zhang.
\newblock Generative adversarial networks as variational training of energy
  based models.
\newblock \emph{arXiv preprint arXiv:1611.01799}, 2016.

\bibitem[Zhang et~al.(2021)Zhang, Xie, Zheng, and Barnes]{zhang2021energy}
J.~Zhang, J.~Xie, Z.~Zheng, and N.~Barnes.
\newblock Energy-based generative cooperative saliency prediction.
\newblock \emph{arXiv preprint arXiv:2106.13389}, 2021.

\bibitem[Zhang et~al.(2017)Zhang, Song, and Qi]{zhang2017age}
Z.~Zhang, Y.~Song, and H.~Qi.
\newblock Age progression/regression by conditional adversarial autoencoder.
\newblock In \emph{Proceedings of the IEEE Conference on Computer Vision and
  Pattern Recognition (CVPR)}, pages 5810--5818, 2017.
\newblock URL \url{https://susanqq.github.io/UTKFace/}.

\bibitem[Zhou et~al.(2020)Zhou, Gao, and Asfour]{zhou2020movement}
Y.~Zhou, J.~Gao, and T.~Asfour.
\newblock Movement primitive learning and generalization: Using mixture density
  networks.
\newblock \emph{IEEE Robotics \& Automation Magazine}, 27\penalty0
  (2):\penalty0 22--32, 2020.

\end{thebibliography}


\clearpage
\appendix

\thispagestyle{empty}

\renewcommand{\thefigure}{S\arabic{figure}}
\setcounter{figure}{0}

\renewcommand{\thetable}{S\arabic{table}}
\setcounter{table}{0}

\renewcommand{\thealgorithm}{S\arabic{algorithm}}
\setcounter{algorithm}{0}

\renewcommand{\theequation}{S\arabic{equation}}
\setcounter{algorithm}{0}

\onecolumn \makesupplementtitle

In this supplementary material, we provide additional details and results. It consists of Appendix~\ref{appendix:societal_impacts} - Appendix~\ref{appendix:code_train}. After discussing limitations and societal impacts in Appendix~\ref{appendix:societal_impacts}, we provide implementation details in Appendix~\ref{appendix:implementation_details}. Then, we describe all utilized datasets more closely in Appendix~\ref{appendix:datasets}. We then provide a complete derivation of Result 1 in Appendix~\ref{appendix:derivation}. Additional results for the 1D regression task is then provided in Appendix~\ref{appendix:results}. Lastly, Appendix~\ref{appendix:code_model} and Appendix~\ref{appendix:code_train} contain example model and training code. Note that figures in this supplementary material are numbered with the prefix "S". Numbers without this prefix refer to the main paper.

\section{LIMITATIONS \& SOCIETAL IMPACTS}
\label{appendix:societal_impacts}

Our approach is primarily intended for regression tasks, where the target space has a limited number of dimensions. For each training sample, several target values are sampled from the proposal distribution. Our approach is therefore not intended to scale to very high-dimensional generative modeling tasks, such as image generation.

Training an EBM using our proposed method in Section~3.2 is somewhat slower than using the NCE baseline method, since we also have to update an MDN proposal at each iteration. The NCE baseline however requires hyperparameters to be tuned specifically for each task at hand. The \emph{total} environmental impact due to training is therefore likely smaller for our proposed method. Our proposed approach for training MDNs in Section~3.3 does however not offer similar benefits compared to conventional MDN training, and is twice as slow to train. This issue would be mitigated to a certain extent by sharing parts of the network among the EBM and MDN, which could be explored in future work. 
\section{IMPLEMENTATION DETAILS}
\label{appendix:implementation_details}

We train all networks for $75$ epochs with a batch size of $32$. The number of samples $M$ is always set to $M = 1024$. All networks are trained on individual NVIDIA TITAN Xp GPUs. Training $20$ networks for a specific setting and dataset on one such GPU takes at most $24-48$ hours. Producing the results in Table 1 to Table 6 thus required approximately $50$ GPU days of training. We utilized an internal GPU cluster.

PyTorch code defining the network architecture used for the head-pose estimation task in Section 5.1 is found in Appendix~\ref{appendix:code_model} below. PyTorch code for the corresponding main training loop is found in Appendix~\ref{appendix:code_train}.

In Section 5.1, the EBM $p(y | x; \theta) = e^{f_\theta(x, y)}/\int e^{f_\theta(x, \tilde{y})} d\tilde{y}$ is evaluated by approximately computing its test set negative log-likelihood (NLL). We do so by evaluating $f_\theta(x, y)$ at densely sampled $y$ values in an interval $[y_{\text{min}}, y_{\text{max}}]$. For the second 1D regression dataset, we evaluate at $8\thinspace192$ values in $[-12.5, 12.5]$. For steering angle prediction, $20\thinspace000$ values in $[-100, 100]$. For cell-count prediction, $19\thinspace900$ values in $[1, 200]$. For age estimation, $5\thinspace900$ values in $[1, 60]$. For head-pose estimation, $27\thinspace000$ values in $\{x \in \mathbb{R}^3: x_i \in [-80, 80], i = 1,2,3\}$.

\section{DATASET DETAILS}
\label{appendix:datasets}

\begin{figure}[t]%
    \begin{minipage}{0.495\textwidth}%
        \centering
        \includegraphics[width=0.825\textwidth]{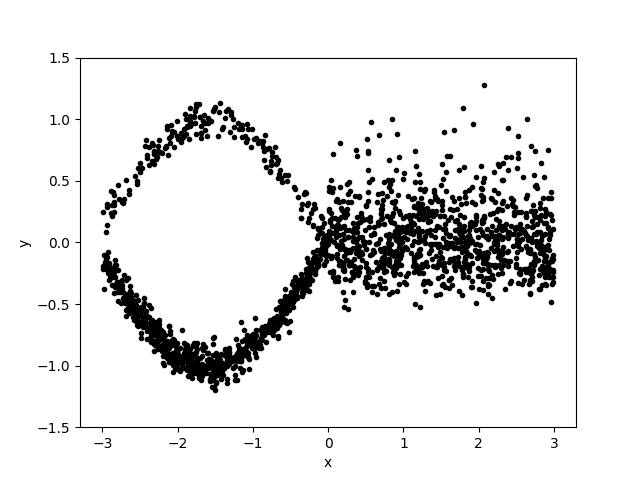}
        \caption{Training data $\{(x_i, y_i)\}_{i=1}^{2000}$ for the first 1D regression dataset~\citep{gustafsson2020train}.}\vspace{-3mm}
        \label{fig:1dregression_1_data}%
    \end{minipage}%
    \quad
    \begin{minipage}{0.495\textwidth}
        \centering
        \includegraphics[width=0.825\textwidth]{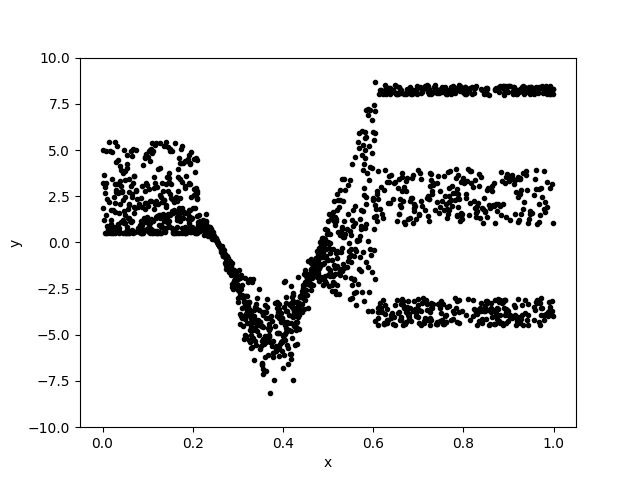}
        \caption{Training data $\{(x_i, y_i)\}_{i=1}^{1700}$ for the second 1D regression dataset~\citep{brando2019modelling}.}\vspace{-3mm}
        \label{fig:1dregression_2_data}%
    \end{minipage}%
\end{figure}

The training data for the two 1D regression problems is visualized in Figure~\ref{fig:1dregression_1_data} and Figure~\ref{fig:1dregression_2_data}.

For steering angle prediction, cell-count prediction and age estimation, our utilized datasets from \citep{ding2021ccgan, ding2020continuous} are all available at \url{https://github.com/UBCDingXin/improved_CcGAN}.

The original age estimation dataset UTKFace~\citep{zhang2017age} is available at \url{https://susanqq.github.io/UTKFace/}, for non-commercial research purposes only. The dataset consists of images collected from the internet, i.e.\ images collected without explicitly obtained consent to be used specifically for training age estimation models. Thus, we choose to not display any dataset examples.

For head-pose estimation, the BIWI~\citep{fanelli2013random} dataset is available for research purposes only. The dataset was created by recording 20 people (research subjects) while they freely turned their heads around. The processed version provided by \citet{yang2019fsa} that we utilize is available at \url{https://github.com/shamangary/FSA-Net}. Since the dataset images could potentially contain personally identifiable information, we choose to not display any dataset examples.

\section{DERIVATION OF RESULT 1}
\label{appendix:derivation}

To derive Result~1 in Section~3.1, we first rewrite $\nabla_{\phi} D_\mathrm{KL}\big(p(y | x; \theta) \parallel q(y | x; \phi)\big)$ according to,
\begin{align*}
    \nabla_{\phi} D_\mathrm{KL}\big(p(y | x; \theta) \parallel q(y | x; \phi)\big) &= \nabla_\phi \int p(y|x;\theta) \log \frac{p(y|x;\theta)}{q(y|x;\phi)} dy\\
    &= \int p(y|x;\theta) \nabla_\phi \log \frac{p(y|x;\theta)}{q(y|x;\phi)} dy\\
    &= \int p(y|x;\theta) \nabla_\phi \bigg(\log p(y|x;\theta) - \log q(y|x;\phi) \bigg) dy\\
    &= - \int p(y|x;\theta) \nabla_\phi \log q(y|x;\phi) dy\\
    &= - \int p(y|x;\theta) \frac{1}{q(y|x;\phi)} \nabla_\phi q(y|x;\phi) dy\\
    &= - \int \frac{e^{f_\theta(x, y)}}{\int e^{f_\theta(x, \tilde{y})} d\tilde{y}} \frac{1}{q(y|x;\phi)} \nabla_\phi q(y|x;\phi) dy\\
    &= - \frac{1}{\int e^{f_\theta(x, \tilde{y})} d\tilde{y}} \int e^{f_\theta(x, y)} \frac{1}{q(y|x;\phi)} \nabla_\phi q(y|x;\phi) dy.
\end{align*}

Then, we approximate the two integrals using Monte Carlo importance sampling,
\begin{align*}
    \nabla_{\phi} D_\mathrm{KL}\big(p(y | x; \theta) \parallel q(y | x; \phi)\big) &= - \frac{1}{\int e^{f_\theta(x, \tilde{y})} d\tilde{y}} \int e^{f_\theta(x, y)} \frac{1}{q(y|x;\phi)} \nabla_\phi q(y|x;\phi) dy\\
    &= - \frac{1}{\int e^{f_\theta(x, y)} dy} \int \frac{e^{f_\theta(x, y)}}{q(y|x;\phi)^2} \big(\nabla_\phi q(y|x;\phi)\big) q(y|x;\phi) dy\\
    &= - \frac{1}{\int \!\frac{e^{f_\theta(x, y)}}{q(y|x;\phi)} q(y|x;\!\phi) dy}\!\int\!\frac{e^{f_\theta(x, y)}}{q(y|x;\!\phi)^2} \big(\nabla_\phi q(y|x;\!\phi)\big) q(y|x;\!\phi) dy\\
    &\approx - \frac{1}{\frac{1}{M}\!\sum_{m=1}^{M}\!\frac{e^{f_\theta(x, y^{(m)})}}{q(y^{(m)}|x;\phi)}} \bigg(\!\frac{1}{M}\!\sum_{m=1}^{M}\!\frac{e^{f_\theta(x, y^{(m)})}}{q(y^{(m)}|x;\phi)^2} \nabla_\phi q(y^{(m)}|x;\phi)\!\bigg),
\end{align*}
where $\{y^{(m)}\}_{m=1}^{M}$ are $M$ independent samples drawn from $q(y | x; \phi)$. Finally, we further rewrite the resulting expression according to,
\begin{align*}
    \nabla_{\phi} D_\mathrm{KL}\big(p(y | x; \theta) \parallel q(y | x; \phi)\big) &\approx - \frac{1}{\frac{1}{M}\!\sum_{m=1}^{M}\!\frac{e^{f_\theta(x, y^{(m)})}}{q(y^{(m)}|x;\phi)}} \bigg(\!\frac{1}{M}\!\sum_{m=1}^{M}\!\frac{e^{f_\theta(x, y^{(m)})}}{q(y^{(m)}|x;\phi)^2} \nabla_\phi q(y^{(m)}|x;\phi) \bigg)\\
    &= \frac{1}{\frac{1}{M}\!\sum_{m=1}^{M}\!\frac{e^{f_\theta(x, y^{(m)})}}{q(y^{(m)}|x;\phi)}} \bigg(\frac{1}{M} \sum_{m=1}^{M} e^{f_\theta(x, y^{(m)})} \nabla_\phi \frac{1}{q(y^{(m)}|x;\phi)} \bigg)\\
    &= \frac{1}{\frac{1}{M}\!\sum_{m=1}^{M}\!\frac{e^{f_\theta(x, y^{(m)})}}{q(y^{(m)}|x;\phi)}} \bigg(\frac{1}{M} \sum_{m=1}^{M} \nabla_\phi \frac{e^{f_\theta(x, y^{(m)})}}{q(y^{(m)}|x;\phi)} \bigg)\\
    &= \frac{1}{\frac{1}{M}\!\sum_{m=1}^{M}\!\frac{e^{f_\theta(x, y^{(m)})}}{q(y^{(m)}|x;\phi)}} \nabla_\phi \bigg(\frac{1}{M} \sum_{m=1}^{M} \frac{e^{f_\theta(x, y^{(m)})}}{q(y^{(m)}|x;\phi)} \bigg)\\
    &= \bigg(\frac{1}{M} \sum_{m=1}^{M} \frac{e^{f_\theta(x, y^{(m)})}}{q(y^{(m)}|x;\phi)}\bigg)^{-1} \nabla_\phi \bigg(\frac{1}{M} \sum_{m=1}^{M} \frac{e^{f_\theta(x, y^{(m)})}}{q(y^{(m)}|x;\phi)} \bigg)\\
    &= \nabla_{\phi} \log \bigg(\frac{1}{M} \sum_{m=1}^{M} \frac{e^{f_{\theta}(x, y^{(m)})}}{q(y^{(m)} | x; \phi)} \bigg).
\end{align*}

\subsection{Best Possible Proposal}
We here expand on the footnote on page 3 of the main paper. When training the EBM $p(y | x; \theta) = e^{f_\theta(x, y)}/Z(x, \theta)$ by minimizing the approximated NLL in (2), we wish to use the proposal $q(y | x; \phi)$ that yields the best possible NLL approximation. In general, this is achieved when the proposal equals the EBM, i.e.\ when $q(y | x; \phi) = p(y | x; \theta)$. To see why this is true, we set $q = p$ in (2),
\begin{align*}
    J(\theta) &= \frac{1}{N} \sum_{i = 1}^{N} \log \bigg(\frac{1}{M} \sum_{m=1}^{M} \frac{e^{f_{\theta}(x_i, y_i^{(m)})}}{q(y_i^{(m)})} \bigg) - f_{\theta}(x_i, y_i)\\
    &= \frac{1}{N} \sum_{i = 1}^{N} \log \bigg(\frac{1}{M} \sum_{m=1}^{M} \frac{e^{f_{\theta}(x_i, y_i^{(m)})}}{p(y_i^{(m)} | x_i; \theta)} \bigg) - f_{\theta}(x_i, y_i)\\
    &= \frac{1}{N} \sum_{i = 1}^{N} \log \bigg(\frac{1}{M} \sum_{m=1}^{M} \frac{e^{f_{\theta}(x_i, y_i^{(m)})}}{e^{f_\theta(x_i, y_i^{(m)})}/Z(x_i, \theta)} \bigg) - f_{\theta}(x_i, y_i)\\
    &= \frac{1}{N} \sum_{i = 1}^{N} \log \bigg(\frac{1}{M} \sum_{m=1}^{M} Z(x_i, \theta) \bigg) - f_{\theta}(x_i, y_i)\\
    &= \frac{1}{N} \sum_{i = 1}^{N} \log Z(x_i, \theta) - f_{\theta}(x_i, y_i)\\
    &= \frac{1}{N} \sum_{i = 1}^{N} - \log \bigg(\frac{e^{f_{\theta}(x_i, y_i)}}{Z(x_i, \theta)}\bigg)\\
    &= \frac{1}{N} \sum_{i = 1}^{N} - \log p(y_i | x_i; \theta),
\end{align*}
which corresponds to the exact NLL objective.

\section{ADDITIONAL RESULTS}
\label{appendix:results}

\begin{figure}
\newcommand{\wid}{3.25cm}%
\newcommand{\imwid}{0.30\textwidth}%
\centering%
			\begin{tabular}{@{\hspace{0.3cm}}c@{\hspace{3.7cm}}c@{\hspace{3.6cm}}c}
				Ground Truth &EBM &MDN Proposal
	\end{tabular}
      \includegraphics*[trim=0 0 0 36, width=\imwid]{figures/1dregression/ground_truth.png}%
      \includegraphics*[trim=0 0 0 36, width=\imwid]{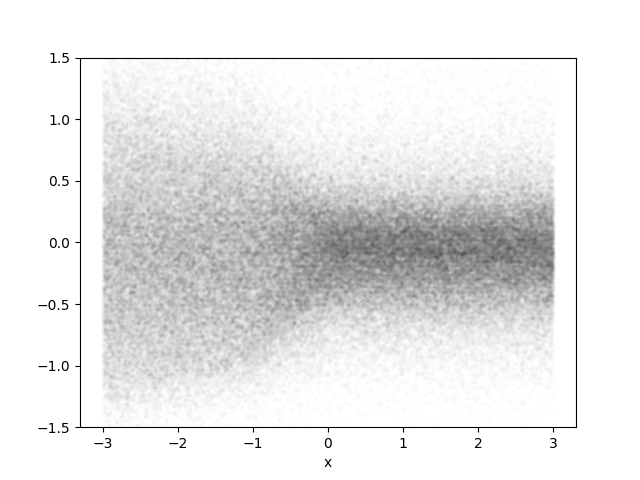}%
      \includegraphics*[trim=0 0 0 36, width=\imwid]{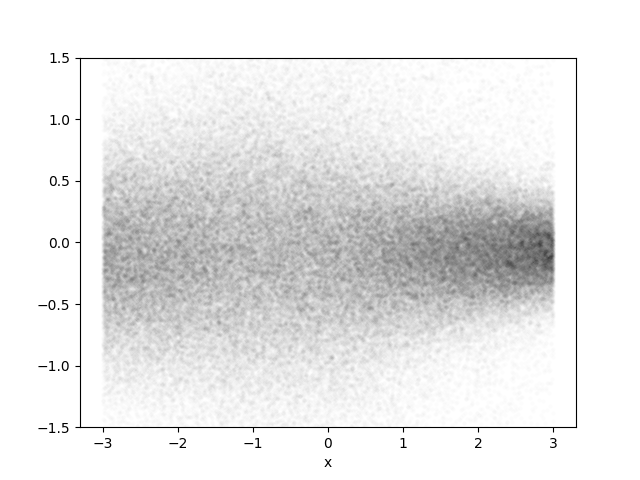}\\
      \includegraphics*[trim=0 0 0 36, width=\imwid]{figures/1dregression/ground_truth.png}%
      \includegraphics*[trim=0 0 0 36, width=\imwid]{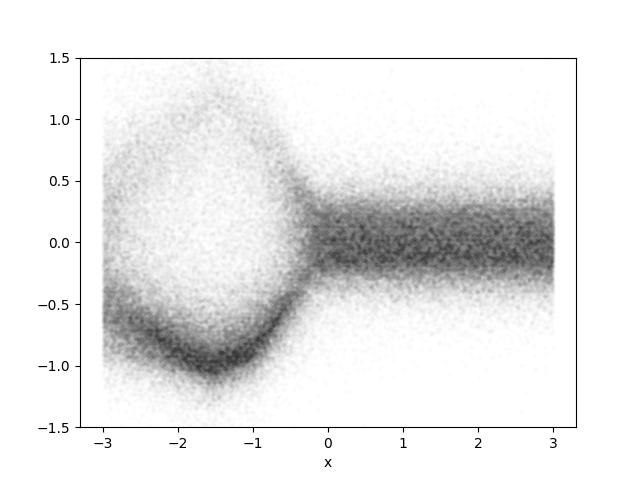}%
      \includegraphics*[trim=0 0 0 36, width=\imwid]{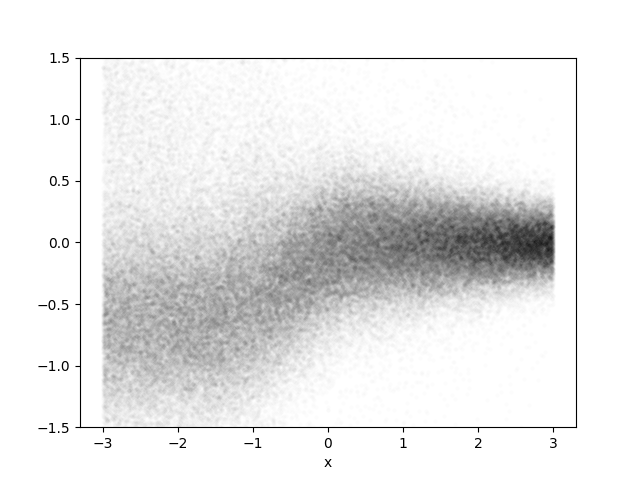}\\
      \includegraphics*[trim=0 0 0 36, width=\imwid]{figures/1dregression/ground_truth.png}%
      \includegraphics*[trim=0 0 0 36, width=\imwid]{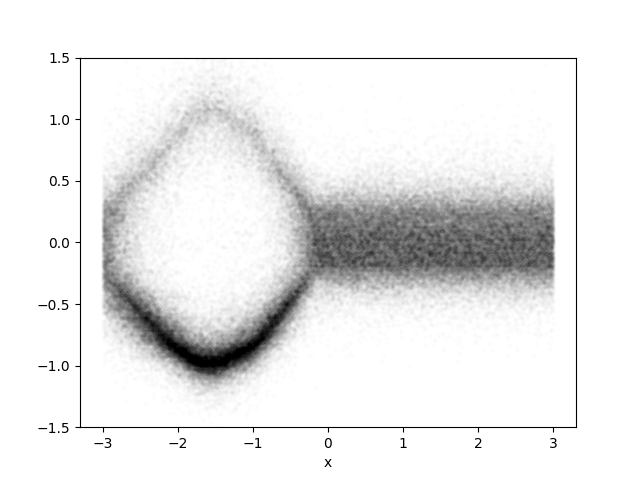}%
      \includegraphics*[trim=0 0 0 36, width=\imwid]{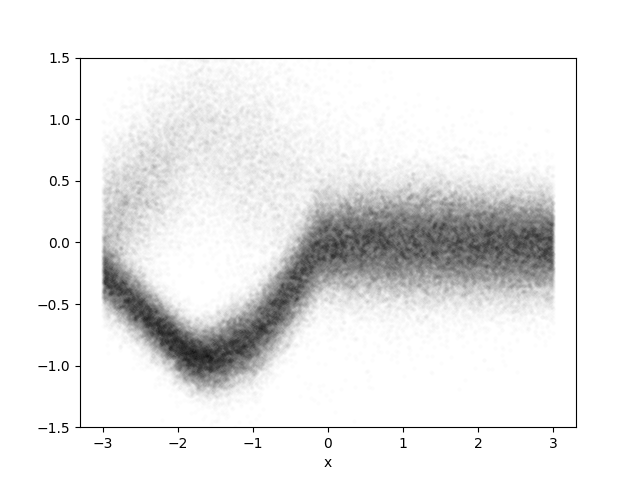}\\
      \includegraphics*[trim=0 0 0 36, width=\imwid]{figures/1dregression/ground_truth.png}%
      \includegraphics*[trim=0 0 0 36, width=\imwid]{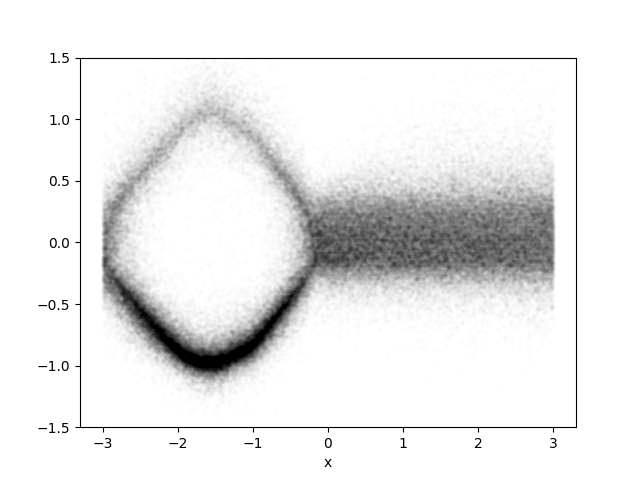}%
      \includegraphics*[trim=0 0 0 36, width=\imwid]{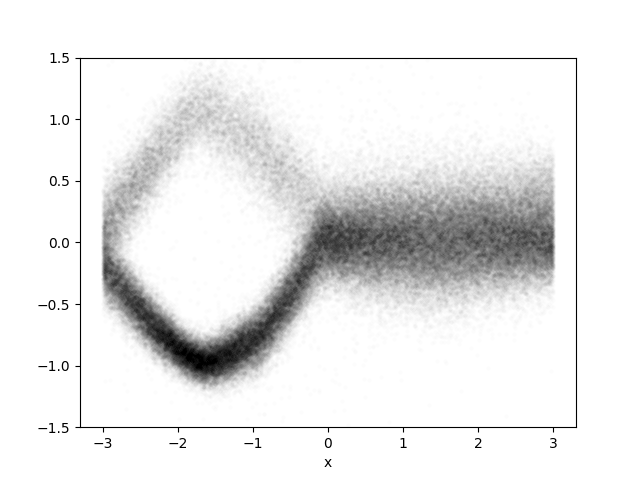}\\
      \includegraphics*[trim=0 0 0 36, width=\imwid]{figures/1dregression/ground_truth.png}%
      \includegraphics*[trim=0 0 0 36, width=\imwid]{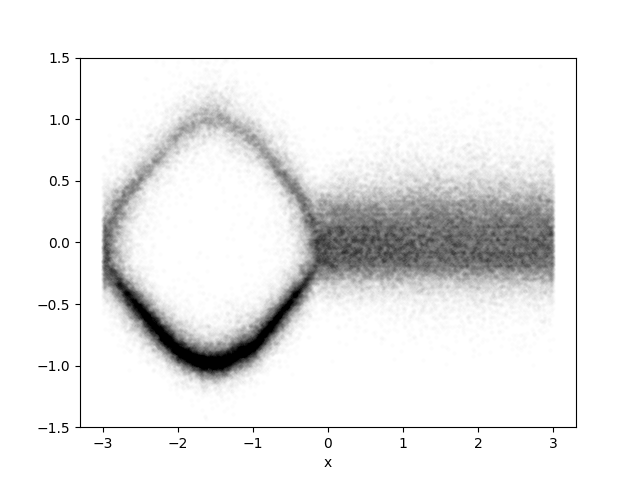}%
      \includegraphics*[trim=0 0 0 36, width=\imwid]{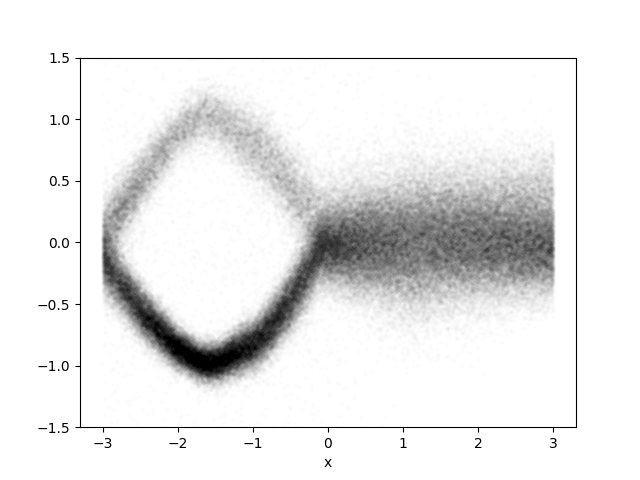}%
\caption{An illustrative 1D regression problem \citep{gustafsson2020train}, demonstrating the effectiveness of our proposed method to jointly train an EBM $p(y | x; \theta)$ and MDN proposal $q(y | x; \phi)$. In this example, the MDN has $K=4$ components. The EBM is trained using NCE with $q(y | x; \phi)$ acting as the noise distribution, whereas the MDN is trained by minimizing its KL divergence to $p(y | x; \theta)$. The EBM and MDN are here visualized after $5$ (top row), $10$, $15$, $20$ and $25$ (bottom row) epochs of training.}\vspace{-3.0mm}
\label{fig:illustrative_extended}%
\end{figure}

Figure 2 in the main paper visualizes the fully trained EBM and MDN proposal, i.e.\ after $75$ epochs of training. In Figure~\ref{fig:illustrative_extended}, we instead visualize the EBM and MDN after $5$ (top row), $10$, $15$, $20$ and $25$ (bottom row) epochs of training. We observe that the EBM is closer to the ground truth early on during training, guiding the MDN via the $J_\mathrm{KL}(\phi)$ loss in (5). 

\begin{figure}
\newcommand{\wid}{3.25cm}%
\newcommand{\imwid}{0.30\textwidth}%
\centering%
			\begin{tabular}{@{\hspace{1.0cm}}c@{\hspace{3.5cm}}c@{\hspace{2.8cm}}c}
				EBM &MDN Proposal &EBM Samples
	\end{tabular}
       \includegraphics*[trim=0 0 0 36, width=\imwid]{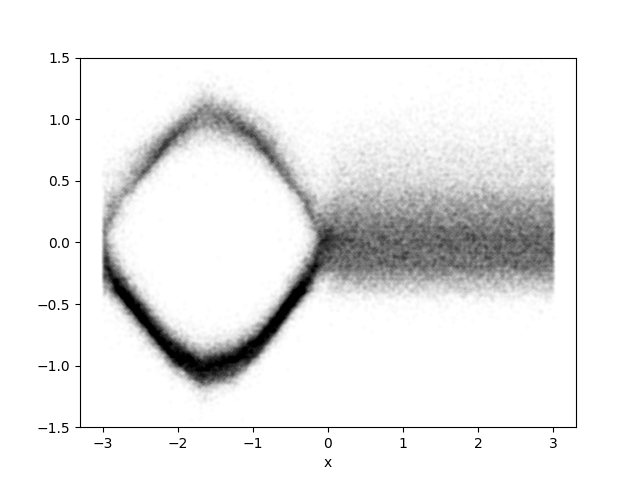}%
       \includegraphics*[trim=0 0 0 36, width=\imwid]{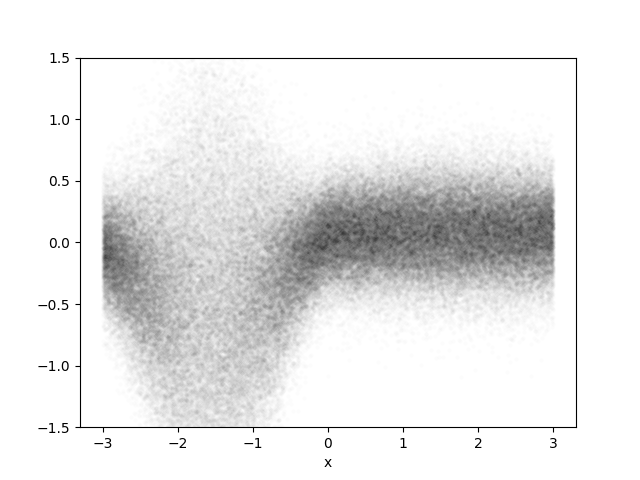}%
       \includegraphics*[trim=0 0 0 36, width=\imwid]{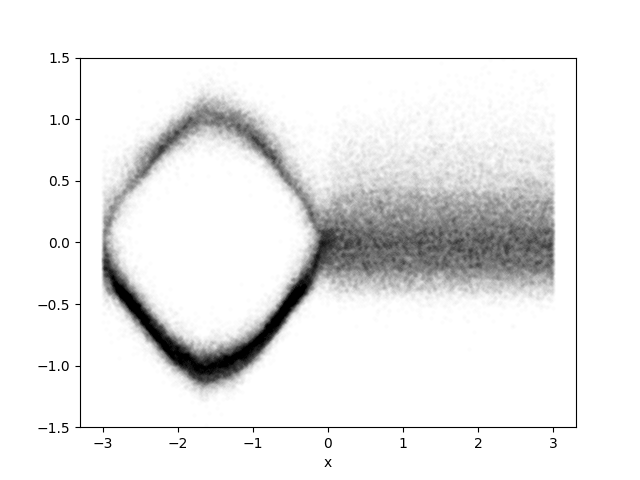}%
\caption{An illustrative 1D regression problem \citep{gustafsson2020train}, demonstrating the effectiveness of our proposed method to jointly train an EBM $p(y | x; \theta)$ and MDN proposal $q(y | x; \phi)$. In this example, the MDN has $K=1$ component. We here also demonstrate that we can draw approximate samples from the EBM using the method described in Section 3.2.2.}
\label{fig:illustrative_sampling}%
\end{figure}

In Figure~\ref{fig:illustrative_sampling}, we visualize the fully trained EBM and MDN proposal when instead using just $K=1$ component in the MDN. We observe that the EBM still is close to the ground truth. Apart from visualizing the EBM using the technique from \citep{gustafsson2020train} (evaluating $f_\theta(x, y)$ at densely sampled $y$ values in the interval $[-3, 3]$ for each $x$), we here also demonstrate that we can draw approximate samples from the EBM using the method described in Section 3.2.2. For each $x$, we draw samples $\{y^{(m)}\}_{m=1}^{1024} \sim q(y | x; \phi)$ from the proposal, compute weights $\{w^{(m)}\}_{m=1}^{1024}$ according to (7), and then re-sample one value from this set $\{y^{(m)}\}_{m=1}^{1024}$ (drawing each $y^{(m)}$ with probability $w^{(m)}$). We observe in Figure~\ref{fig:illustrative_sampling} that this method produces accurate EBM samples, even when the proposal is unimodal and thus not a particularly close approximation of the EBM.  

\section{PYTORCH CODE - NETWORK ARCHITECTURE}
\label{appendix:code_model}

\begin{minted}{python}
class NoiseNet(nn.Module):
    def __init__(self, hidden_dim):
        super().__init__()

        self.K = 4

        self.fc1_mean = nn.Linear(hidden_dim, hidden_dim)
        self.fc2_mean = nn.Linear(hidden_dim, 3*self.K)

        self.fc1_sigma = nn.Linear(hidden_dim, hidden_dim)
        self.fc2_sigma = nn.Linear(hidden_dim, 3*self.K)

        self.fc1_weight = nn.Linear(hidden_dim, hidden_dim)
        self.fc2_weight = nn.Linear(hidden_dim, self.K)

    def forward(self, x_feature):
        means = F.relu(self.fc1_mean(x_feature))
        means = self.fc2_mean(means)
        
        log_sigma2s = F.relu(self.fc1_sigma(x_feature))
        log_sigma2s = self.fc2_sigma(log_sigma2s)

        weight_logits = F.relu(self.fc1_weight(x_feature))
        weight_logits = self.fc2_weight(weight_logits)
        weights = torch.softmax(weight_logits, dim=1)

        return means, log_sigma2s, weights

class PredictorNet(nn.Module):
    def __init__(self, input_dim, hidden_dim):
        super().__init__()

        self.fc1_y = nn.Linear(input_dim, 16)
        self.fc2_y = nn.Linear(16, 32)
        self.fc3_y = nn.Linear(32, 64)
        self.fc4_y = nn.Linear(64, 128)

        self.fc1_xy = nn.Linear(hidden_dim+128, hidden_dim)
        self.fc2_xy = nn.Linear(hidden_dim, 1)

    def forward(self, x_feature, y):
        batch_size, num_samples, _ = y.shape

        x_feature = x_feature.view(batch_size, 1, -1).expand(-1, num_samples, -1)
        x_feature = x_feature.reshape(batch_size*num_samples, -1)
        
        y = y.reshape(batch_size*num_samples, -1)

        y_feature = F.relu(self.fc1_y(y))
        y_feature = F.relu(self.fc2_y(y_feature))
        y_feature = F.relu(self.fc3_y(y_feature))
        y_feature = F.relu(self.fc4_y(y_feature))

        xy_feature = torch.cat([x_feature, y_feature], 1)

        xy_feature = F.relu(self.fc1_xy(xy_feature))
        score = self.fc2_xy(xy_feature)

        score = score.view(batch_size, num_samples)

        return score
        
class FeatureNet(nn.Module):
    def __init__(self):
        super().__init__()

        resnet18 = models.resnet18(pretrained=True)
        self.resnet18 = nn.Sequential(*list(resnet18.children())[:-2])

        self.avg_pool = nn.AdaptiveAvgPool2d((1, 1))

    def forward(self, x):
        x_feature = self.resnet18(x)
        x_feature = self.avg_pool(x_feature)
        x_feature = x_feature.squeeze(2).squeeze(2) 

        return x_feature

class Net(nn.Module):
    def __init__(self):
        super(Net, self).__init__()

        hidden_dim = 512

        self.feature_net = FeatureNet()
        self.noise_net = NoiseNet(hidden_dim)
        self.predictor_net = PredictorNet(3, hidden_dim)

    def forward(self, x, y):
        x_feature = self.feature_net(x)
        return self.noise_net(x_feature)
\end{minted}
\section{PYTORCH CODE - TRAINING LOOP}
\label{appendix:code_train}

\begin{minted}{python}
for step, (xs, ys) in enumerate(train_loader):
    xs = xs.cuda() # (shape: (batch_size, 3, img_size, img_size))
    ys = ys.cuda() # (shape: (batch_size, 3))

    x_features = network.feature_net(xs) # (shape: (batch_size, hidden_dim))

    means, log_sigma2s, weights = network.noise_net(x_features.detach())
    # (means has shape: (batch_size, 3K))
    # (log_sigma2s has shape: (batch_size, 3K))
    # (weights has shape: (batch_size, K))
    sigmas = torch.exp(log_sigma2s/2.0) # (shape: (batch_size, 3K))
    means = means.view(-1, 3, K) # (shape: (batch_size, 3, K))
    sigmas = sigmas.view(-1, 3, K) # (shape: (batch_size, 3, K))

    q_distr = torch.distributions.normal.Normal(loc=means, scale=sigmas)
    q_ys_K = torch.exp(q_distr.log_prob(ys.unsqueeze(2)).sum(1)) # (shape: (batch_size, K)
    q_ys = torch.sum(weights*q_ys_K, dim=1) # (shape: (batch_size))

    y_samples_K = q_distr.sample(sample_shape=torch.Size([num_samples])) 
    # (shape: (num_samples, batch_size, 3, K))
    inds = torch.multinomial(weights, num_samples=num_samples,
                             replacement=True).unsqueeze(2).unsqueeze(2) 
    # (shape: (batch_size, num_samples, 1, 1))
    inds = inds.expand(-1, -1, 3, 1) # (shape: (batch_size, num_samples, 3, 1))
    inds = torch.transpose(inds, 1, 0) # (shape: (num_samples, batch_size, 3, 1))
    y_samples = y_samples_K.gather(3, inds).squeeze(3) # (shape: (num_samples, batch_size, 3))
    y_samples = y_samples.detach()
    q_y_samples_K = torch.exp(q_distr.log_prob(y_samples.unsqueeze(3)).sum(2)) 
    # (shape: (num_samples, batch_size, K))
    q_y_samples = torch.sum(weights.unsqueeze(0)*q_y_samples_K, dim=2) 
    # (shape: (num_samples, batch_size))
    y_samples = torch.transpose(y_samples, 1, 0) # (shape: (batch_size, num_samples, 3))
    q_y_samples = torch.transpose(q_y_samples, 1, 0) # (shape: (batch_size, num_samples))

    scores_gt = network.predictor_net(x_features, ys.unsqueeze(1)) # (shape: (batch_size, 1))
    scores_gt = scores_gt.squeeze(1) # (shape: (batch_size))

    scores_samples = network.predictor_net(x_features, y_samples) 
    # (shape: (batch_size, num_samples))

    ########################################################################
    # compute loss:
    ########################################################################
    f_samples = scores_samples
    p_N_samples = q_y_samples.detach()
    f_0 = scores_gt
    p_N_0 = q_ys.detach()
    exp_vals_0 = f_0-torch.log(p_N_0)
    exp_vals_samples = f_samples-torch.log(p_N_samples)
    exp_vals = torch.cat([exp_vals_0.unsqueeze(1), exp_vals_samples], dim=1)
    loss_ebm_nce = -torch.mean(exp_vals_0 - torch.logsumexp(exp_vals, dim=1))

    log_Z = torch.logsumexp(scores_samples.detach() 
                            - torch.log(q_y_samples), dim=1) - math.log(num_samples) 
    loss_mdn_kl = torch.mean(log_Z)

    loss = loss_ebm_nce + loss_mdn_kl

    optimizer.zero_grad()
    loss.backward()
    optimizer.step()
\end{minted}

\end{document}